\documentclass{article} %
\usepackage{iclr2027_conference,times}

\usepackage{amsmath,amsfonts,bm}

\def\eqref#1{equation~\ref{#1}}

\def\1{\bm{1}}

\DeclareMathAlphabet{\mathsfit}{\encodingdefault}{\sfdefault}{m}{sl}
\SetMathAlphabet{\mathsfit}{bold}{\encodingdefault}{\sfdefault}{bx}{n}

\usepackage{xspace}
\usepackage{multirow}
\usepackage{lipsum}
\usepackage{colortbl}
\usepackage{graphicx}
\usepackage{siunitx}
\usepackage{duckuments}
\usepackage{pifont} %
\usepackage{bm}
\usepackage{booktabs}
\usepackage{hyperref}
\usepackage{url}
\usepackage{cleveref}
\usepackage[]{mdframed}

\usepackage{algorithmic}
\usepackage{algorithm}

\definecolor{LightGray}{gray}{1}
\definecolor{lightgray}{gray}{0.95}
\definecolor{midgray}{gray}{0.55}
\definecolor{steelblue}{HTML}{4D82B7}
\definecolor{davysgrey}{rgb}{0.33, 0.33, 0.33}
\definecolor{LightCyan}{rgb}{0.88,1,1}
\definecolor{LightGold}{HTML}{F3E2C5}
\definecolor{AngelRow}{HTML}{FFFDD0}
\definecolor{ao(english)}{rgb}{0.0, 0.5, 0.0}
\definecolor{lightsalmon}{rgb}{1.0, 0.63, 0.48}
\definecolor{tabgreen}{HTML}{2CA02C}
\definecolor{newgreen}{HTML}{CCE0AC}
\definecolor{newblue}{HTML}{6bc5ff}
\definecolor{viola}{HTML}{935fa7}
\newcommand{\ourcolor}{viola!20}

\crefname{section}{Sec.}{Secs.}
\crefname{table}{Tab.}{Tabs.}
\crefname{figure}{Fig.}{Figs.}

\Crefname{section}{Section}{Sections}
\Crefname{table}{Table}{Tables}

\newcommand{\theseus}{\textsc{Theseus}\xspace}
\newcommand{\bico}{BiCo\xspace}

\newcommand{\red}[1]{\textcolor{red}{#1}}

\newcommand{\dgreen}[1]{\textcolor{ao(english)}{#1}}

\newcommand{\cmark}{\ding{51}}%
\newcommand{\greencmark}{\dgreen{\cmark}}
\newcommand{\xmark}{\ding{55}}%
\newcommand{\redxmark}{\red{\xmark}}

\newcommand{\tit}[1]{\smallbreak\noindent\textbf{#1 }}

\crefname{section}{Sec.}{Secs.}
\crefname{table}{Tab.}{Tabs.}
\crefname{figure}{Fig.}{Figs.}
\crefname{equation}{Eq.}{Eqs.}

\newcommand{\ourrow}{\rowcolor{\ourcolor}[\dimexpr\tabcolsep+0.1pt\relax]}
\newcommand{\grayrow}{\rowcolor{gray!20}[\dimexpr\tabcolsep+0.1pt\relax]}

\makeatletter
\DeclareRobustCommand\onedot{\futurelet\@let@token\@onedot}
\def\@onedot{\ifx\@let@token.\else.\null\fi\xspace}

\def\ie{\emph{i.e}\onedot}

\makeatother

\usepackage{array}
\newcommand{\PreserveBackslash}[1]{\let\temp=\\#1\let\\=\temp}
\newcolumntype{C}[1]{>{\PreserveBackslash\centering}p{#1}}
\newcolumntype{R}[1]{>{\PreserveBackslash\raggedleft}p{#1}}
\newcolumntype{L}[1]{>{\PreserveBackslash\raggedright}p{#1}}
\newcolumntype{Y}{>{\centering\arraybackslash}X}

\newcommand{\plus}{\texttt{+}}
\newcommand{\minus}{\texttt{-}}

\newcommand{\deltaneg}[1]{\textcolor{red}{\scriptsize{(\minus#1)}}}
\newcommand{\deltapos}[1]{\textcolor{green!60!black}{\scriptsize{(\plus#1)}}}
\newcommand{\deltazero}[0]{\textcolor{black}{\scriptsize{(\plus0.00)}}}

\definecolor{MaterialRed}{HTML}{D32F2F}        %
\definecolor{MaterialBlue}{HTML}{1976D2}       %
\definecolor{MaterialGreen}{HTML}{388E3C}      %
\definecolor{MaterialOrange}{HTML}{F57C00}     %
\definecolor{MaterialPurple}{HTML}{7B1FA2}     %
\definecolor{MaterialTeal}{HTML}{00796B}       %
\definecolor{MaterialIndigo}{HTML}{303F9F}     %
\definecolor{MaterialBrown}{HTML}{5D4037}      %

\definecolor{srcact}{RGB}{121,181,211}   %
\definecolor{tgtact}{RGB}{200,149,89}    %
\definecolor{tgtparam}{RGB}{93,153,51}   %
\definecolor{ftpink}{RGB}{223,161,182}   %

\title{From Parameter Rebasin to Functional Transfer Across Heterogeneous Transformers}

\author{%
\textbf{Filippo Rinaldi$^1$}, \textbf{Aniello Panariello$^1$}, \textbf{Giacomo Salici$^1$}, \textbf{Simone Magistri$^2$},\\
\textbf{Angelo Porrello$^1$}, \textbf{Andrew D. Bagdanov$^2$}, \textbf{Simone Calderara$^1$} \vspace{.5em}\\
$^1$University of Modena and Reggio Emilia, Italy \\
$^2$University of Florence, Italy \\[0.4em]
\texttt{name.surname@unimore.it, name.surname@unifi.it}%
}

\newcommand{\method}{\textsc{Ariadne}}
\newcommand{\methodgrad}{\textsc{Ariadne}\ensuremath{_{\nabla}}}

\iclrfinalcopy %
\begin{document}

\maketitle

\begin{abstract}
Adapting pretrained Transformers to downstream tasks produces task-specific updates that are expensive to reproduce for every model variant. Reusing these updates across different depths and widths is challenging because their parameters have incompatible shapes and do not share a canonical coordinate system. We introduce \method{}, a training-free method that transfers task-specific behavior across heterogeneous Transformers without resizing the source model or transporting its parameters. Rather than treating the source task vector, \ie, the difference in parameter space between the fine-tuned and the base models, as the object to move, \method{} characterizes fine-tuning by the activation changes it induces at matched blocks. First, it aligns these effects to a target base through local Procrustes maps. Then, it directly synthesizes a native target update by closed-form ridge regression. Across Transformer expansion, reduction and language-model transfer, \method{} improves target-model performance while substantially reducing the time and memory required to construct an update relative to previous cross-architecture rebasin approaches. These results show that local functional effects provide an efficient alternative to parameter transport.
\end{abstract}

\section{Introduction}
Foundation models~\citep{radford2021learning,yang2025qwen3} are increasingly reused rather than trained from scratch. A single pretrained checkpoint can give rise to many fine-tuned task variants,  represented as task vectors~\citep{ilharco2023editing}, each defined in the parameterization of its source architecture. When a deployment requires a different backbone, the learned task-specific behavior cannot be directly transferred because source and target models inhabit different parameter spaces. Valuable model families consequently remain tied to the architecture on which they were trained.

This coupling is limiting in practice. For example, a practitioner may have an instruction-tuned assistant at one Transformer depth and width, but later need to deploy the same instruction-following behavior in a smaller model under a tighter serving budget, or in a newer and larger backbone. Similarly, reducing depth can lower latency and memory use, while expansion can make an existing model family compatible with a larger target architecture. Retraining every task-specific model in the new architecture may be unavailable or prohibitively expensive; it also repeats substantial training compute and its associated energy use. The practical goal is therefore to reuse task-specific behavior across heterogeneous Transformer architectures without repeating fine-tuning.

Existing work tackles this challenge by either adapting pretrained architectures or transporting task-specific updates across models. Function-preserving transformations such as Net2Net~\citep{chen2015net2net} and Network Morphism~\citep{wei2016network} expand trained networks, while Transformer-specific growth and compression methods often rely on parameter transformations or continued training~\citep{gesmundo2023composable,ding2023network,chen2022bert2bert,fan2020Reducing,sajjad2023effect,xu2020bert}. A natural alternative is to align source and target parameterizations and then transport the source task vector, as in cross-architecture rebasin methods such as \theseus{}~\citep{rinaldi2026transporting} and \bico{}~\citep{son2026bico}. This formulation treats the source task vector, namely the parameter change caused by fine-tuning, as the object to move. 

With reference to these transport-based methods~\citep{rinaldi2026transporting,son2026bico}, a sufficiently accurate correspondence in parameter space or at the global functional level is therefore required. This typically entails assumptions on the backbone and relative depth of the models, together with architectural resampling operations such as block copying, interpolation, or deletion to match their depths. These choices are inherently heuristic and do not guarantee that the transported task vector remains informative. More elaborate resampling procedures can mitigate this issue~\citep{kangaslahti2026boomerang}, but require gradient access. In this work, we therefore seek transfer techniques that are simple, efficient, and robust to mismatches between model parameterizations.

To reach these goals, \method{} adopts a different transfer paradigm. Rather than transporting the source task vector, it transfers the observable effect of source fine-tuning, that is the change in the transformer block activation induced by fine-tuning. For each matched pair of Transformer blocks, \method{} aligns source and target base residual-stream states with a local Procrustes map and uses it to express the source fine-tuning activation displacement in target coordinates. Then, a closed-form ridge regression computes an additive update to one residual-writing target component that matches this desired effect. The base-model relation is therefore only a local coordinate bridge, not the optimization target. The resulting update is native to the target backbone and requires neither a target fine-tuned checkpoint, source resizing, parameter transport, nor inference-time interventions. The name of the method recalls Ariadne's thread: the activation displacement guides construction of an update through a differently structured target model. Our contributions are:
\begin{itemize}
  \item We formulate transfer across Transformer depths and widths as direct reproduction of fine-tuning-induced functional effects, rather than parameter rebasin.
  \item We introduce a training-free local Procrustes and ridge construction that synthesizes target-space task vectors from a source base and fine-tuned counterpart.
  \item We evaluate direct functional transfer across Vision Transformer expansion and reduction, and across language-model variants.
\end{itemize}

\section{Related Work}

\tit{Rebasin and cross-architecture transfer.} Rebasin methods align parameterizations through permutation symmetries, enabling task-vector reuse across pretrained checkpoints~\citep{ainsworth2023git,rinaldi2025update}. GradFix~\citep{rinaldi2026gradient} transfers updates between models with the same architecture by masking source task vectors using target gradient signs from a small labeled calibration set. \theseus{}~\citep{rinaldi2026transporting} and \bico{}~\citep{son2026bico} combine architectural alignment with Procrustes-based task-vector transport across different widths and depths. \theseus{} aligns input and output activations, while \bico{} replaces output activations with output-side gradients. \method{} instead fits a native target update to the aligned activation difference induced by source fine-tuning, without transporting source tensors.

\tit{Representation alignment and activation matching.} Model stitching connects subnetworks through mappings between intermediate representations~\citep{bansal2021revisiting}. Relative representations use similarities to shared anchors~\citep{moschella2023relative}, while latent-space translation estimates algebraic maps between pretrained feature spaces~\citep{maiorca2023latent}. Representation metrics further formalize comparisons under transformations of hidden coordinates~\citep{williams2021generalized}. Activation matching also provides an objective for constructing model weights. RegMean fits merged linear layers by minimizing output discrepancies with the original models~\citep{jin2023regmean}; PLeaS combines partial permutation matching with layer-wise least squares to reconstruct features at a chosen model size~\citep{nasery2025pleas}.

\tit{Distillation and network transformation.} Knowledge distillation transfers knowledge from a teacher model to a student by training the latter to reproduce the teacher's predictions~\citep{hinton2015distilling}. This allows learned behavior to be reused across models of different sizes without requiring a correspondence between their parameters. TinyBERT extends this supervision to Transformer representations at both general and task-specific stages~\citep{jiao2020tinybert}. Architectural transformations offer another route to model reuse: Net2Net~\citep{chen2015net2net} and Network Morphism~\citep{wei2016network} preserve network functions during architectural changes, while Transformer growth and compression methods often use continued training or layer removal~\citep{gesmundo2023composable,ding2023network,chen2022bert2bert,fan2020Reducing,sajjad2023effect,xu2020bert}. Boomerang Distillation reinserts teacher blocks into a distilled student to obtain intermediate model sizes~\citep{kangaslahti2026boomerang}.

\begin{figure}
    \centering
    \includegraphics[width=.95\linewidth]{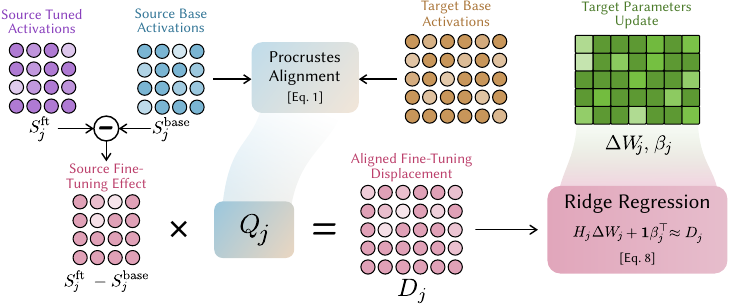}
   \caption{\textbf{\method{} transfers the functional effect of source fine-tuning to the target model in activation space.} Source \textcolor{srcact}{\ding{108}} and target \textcolor{tgtact}{\ding{108}} base activations are aligned using a component-wise Procrustes transformation $Q_{j}$. The source fine-tuning effect \textcolor{ftpink}{\ding{108}} is  mapped into target coordinates to obtain the desired activation shift $D_{j}$, giving the aligned fine-tuning displacement. Ridge regression converts this desired shift into target-native parameter updates \textcolor{tgtparam}{\ding{110}}.}
    \label{fig:main}
\end{figure}
\section{Method}
\label{sec:method}
\method{}, depicted in~\cref{fig:main}, transfers task-specific behavior directly to a native target backbone. Let $\theta_{\mathrm{s}}^{\mathrm{base}}$ and $\theta_{\mathrm{s}}^{\mathrm{ft}}$ be a source base checkpoint and its fine-tuned counterpart, with source task vector $\tau_{\mathrm{s}}=\theta_{\mathrm{s}}^{\mathrm{ft}}-\theta_{\mathrm{s}}^{\mathrm{base}}$. Given an untouched target base $\theta_{\mathrm{t}}^{\mathrm{base}}$, possibly with different depth, width, or block design, we construct a sparse target update $\tau_{\mathrm{t}}$ whose selected residual-stream effects match the source fine-tuning effects on a small calibration set. Crucially, $\tau_{\mathrm{t}}$ is not obtained by transporting $\tau_{\mathrm{s}}$ into the target parameter space. Instead, \method{} synthesizes $\tau_{\mathrm{t}}$ directly in the target parameterization to reproduce the functional change induced by moving from $\theta_{\mathrm{s}}^{\mathrm{base}}$ to $\theta_{\mathrm{s}}^{\mathrm{ft}}$. The method neither resizes source checkpoints nor seeks functional equivalence of the source and target base models.
\subsection{Block-level Functional Correspondence}
Prior rebasin methods transfer task vectors in parameter space: given a source task vector, they transform its parameter update and inject it into the target model. \method{} moves this problem to activation space. Rather than asking how to transform $\tau_{\mathrm{s}}$, we characterize the activation change induced by fine-tuning and express that change in target coordinates.

We use the term \emph{block} to denote a recurring computational unit of the backbone, typically comprising an aggregation of operations such as self-attention and feed-forward transformations, as commonly found in
Transformer architectures. When transferring information between source and target architectures, we therefore require a correspondence between their blocks, specifying which source blocks provide the information to be transferred to which target blocks. Let $\pi$ be a user-specified, possibly partial map from target to source blocks, where $\pi(j)$ identifies the source block whose fine-tuning effect supervises target block $j$. We use relative-depth pairing, $\pi(j)=\operatorname{round}\!\left(j(D_{\mathrm{s}}-1)/(D_{\mathrm{t}}-1)\right)$, although structural ancestry or domain knowledge may also define the map. The map $\pi$ does not need to be one-to-one; indeed, when expansion may reuse source blocks, while reduction may omit some. 

For every $j\in\operatorname{dom}(\pi)$, we extract the output of target block $j$ from the pre-trained target model, denoted by $T_j^{\mathrm{base}}$, and the output of the mapped source block $\pi(j)$ from both the pre-trained and fine-tuned source models, denoted by $S_{\pi(j)}^{\mathrm{base}}$ and $S_{\pi(j)}^{\mathrm{ft}}$. We use a small calibration set of $n$ examples to compute these activation
banks, which are centered around zero by subtracting their mean. When source
and target token grids differ, we additionally interpolate the source
activations to the target grid before row-wise comparison,
following~\cite{rinaldi2026transporting}.

Next, we align the two base activations spaces by solving the rectangular Procrustes problem:
\begin{equation}
Q_j=\arg\min_{R\in\mathcal P_{d_{\mathrm{s}},d_{\mathrm{t}}}}
\left\lVert\widetilde S_{\pi(j)}^{\mathrm{base}}R-\widetilde T_j^{\mathrm{base}}\right\rVert_F^2,
\qquad
Q_j\in\mathbb{R}^{d_{\mathrm{s}}\times d_{\mathrm{t}}},
\label{eq:procrustes_objective}
\end{equation}
where $\mathcal P_{d_{\mathrm{s}},d_{\mathrm{t}}}$ denotes the appropriate rectangular Stiefel manifold, \ie, matrices with orthonormal columns when $d_{\mathrm{s}}\geq d_{\mathrm{t}}$ and orthonormal rows otherwise. With the thin SVD
\begin{equation}
Q_j=U_jV_j^{\top} \qquad \text{where} \qquad \left(\widetilde S_{\pi(j)}^{\mathrm{base}}\right)^{\top}\widetilde T_j^{\mathrm{base}}=U_j\Sigma_jV_j^{\top}
\qquad
.
\label{eq:procrustes}
\end{equation}
We then apply the learned map $Q_j$ to the source fine-tuning effect, defined as the activation displacement $S_{\pi(j)}^{\mathrm{ft}}-S_{\pi(j)}^{\mathrm{base}}$. This expresses the source fine-tuning effect in the target coordinates:
\begin{equation}
D_j=\left(S_{\pi(j)}^{\mathrm{ft}}-S_{\pi(j)}^{\mathrm{base}}\right)Q_j,
\qquad D_j\in\mathbb{R}^{n\times d_{\mathrm{t}}}.
\label{eq:desired_effect}
\end{equation}
The quantity $D_j$ is the desired residual-stream displacement after target
block $j$. It plays the role that a transported task vector would play in
parameter rebasin, but in activation space: it specifies the effect that
the target component must realize in the residual stream. The regression
in the next subsection turns this effect into the additive target parameters
$\Delta W_j$ and $\beta_j$ that form $\tau_{\mathrm{t}}$.

\begin{algorithm}[t]
\caption{\method{}}
\label{alg:direct-functional-transfer}
\small
\begin{algorithmic}[1]
\REQUIRE Source models $\theta_{\mathrm{s}}^{\mathrm{base}}$ and
$\theta_{\mathrm{s}}^{\mathrm{ft}}$, target base
$\theta_{\mathrm{t}}^{\mathrm{base}}$, calibration set, block map $\pi$
\ENSURE Target task vector $\tau_{\mathrm{t}}$

\STATE Run the source base, source fine-tuned, and target base models on the
calibration set.
\STATE Collect $S_{\pi(j)}^{\mathrm{base}}$,
$S_{\pi(j)}^{\mathrm{ft}}$, $T_j^{\mathrm{base}}$, and $H_j$ for all
$j\in\operatorname{dom}(\pi)$.
\STATE Initialize $\tau_{\mathrm{t}}\leftarrow 0$.

\FOR{each $j\in\operatorname{dom}(\pi)$}
    \STATE Compute the source-to-target alignment $Q_j$ using
    \cref{eq:procrustes_objective}.
    \STATE Transport the source fine-tuning effect:
    $D_j\leftarrow
    (S_{\pi(j)}^{\mathrm{ft}}-S_{\pi(j)}^{\mathrm{base}})Q_j$.
    \STATE Compute the ridge statistics from
    \cref{eq:direct_ridge} and select $\lambda_j$ using
    \cref{eq:empirical_bayes_inverse}.
    \STATE Solve \cref{eq:direct_ridge_solution} for
    $(\Delta W_j,\beta_j)$ and insert the update into $\tau_{\mathrm{t}}$.
\ENDFOR

\STATE \textbf{return} $\tau_{\mathrm{t}}$
\end{algorithmic}
\end{algorithm}

\tit{Invariance to source activation coordinates.} The columns of a collected source residual-stream matrix are hidden feature coordinates, whose orientation is not intrinsic. Consider an equivalent source representation obtained by applying an orthogonal change of basis $O\in\mathbb{R}^{d_{\mathrm{s}}\times d_{\mathrm{s}}}$ to every source residual vector. The collected base and fine-tuned state matrices then become $S_{\pi(j)}^{\mathrm{base}}O$ and $S_{\pi(j)}^{\mathrm{ft}}O$. If $Q_j$ aligns the original source-base states to the target-base states, a corresponding optimal Procrustes bridge after this change of coordinates is $Q'_j=O^\top Q_j$, since
\begin{equation}
\left(\widetilde S_{\pi(j)}^{\mathrm{base}}O\right)Q'_j
=
\widetilde S_{\pi(j)}^{\mathrm{base}}Q_j.
\end{equation}
The aligned fine-tuning displacement is therefore also unchanged:
\begin{equation}
D'_j
=
\left(
S_{\pi(j)}^{\mathrm{ft}}O-S_{\pi(j)}^{\mathrm{base}}O
\right)Q'_j
=
\left(
S_{\pi(j)}^{\mathrm{ft}}-S_{\pi(j)}^{\mathrm{base}}
\right)Q_j
=
D_j.
\end{equation}
Thus, the desired target effect is independent of the particular orthogonal coordinates used to represent the source hidden states. In contrast, a parameter update changes under such a reparameterization, because the affected source weights must change with the hidden coordinates. Recovering these parameter-coordinate transformations across models with different widths, depths, or block structures is precisely the difficult part of rebasin. \method{} instead transfers a local activation effect whose target-space specification is unchanged by this coordinate choice.

\tit{Alternative alignment signals: \methodgrad.} Our default constructs $Q_j$ from source and target base residual-stream states. More generally, the alignment can use any paired source--target signal whose source coordinates transform as $X_{\mathrm{s}}\mapsto X_{\mathrm{s}}O$ under an orthogonal change of source basis. Following \bico, we devise an augmented variant, referred to as \methodgrad, which constructs $Q_j$ from gradients of the task loss with respect to each block output. These row-vector gradients obey the same transformation rule, so the invariance argument above and the resulting activation displacement $D_j$ remain unchanged. Unlike activation alignment, this label-dependent variant requires backward passes, we therefore treat it as an alternative approach rather than the default method.
\subsection{Direct Component Regression}
The desired residual-stream displacement $D_j$ specifies how the target representation should change on the calibration inputs, but is not itself deployable: it is an input-dependent activation matrix rather than a target-model parameter update. We instead seek a task vector that can be added once to the target checkpoint and used on new inputs without activation interventions. To this end, we will absorb the effect encoded in $D_j$ into the parameters of a selected component of a block. In our experiments, this component corresponds to \texttt{c\_proj} for CLIP-based Vision Transformers and \texttt{down\_proj} for LLMs in the Qwen family, \ie, the final MLP projection before the residual addition.

For each $j\in\operatorname{dom}(\pi)$, let $H_j\in\mathbb{R}^{n\times d^{\mathrm{in}}_j}$ denote the activation matrix entering the selected target component in block $j$ with $n$ samples. Let its original output be $H_jW_j+\mathbf{1}b_j^\top$. We seek an additive update to its existing parameters that compiles the effect encoded in $D_j$:
\begin{equation}
W_j' = W_j+\Delta W_j,
\qquad
b_j' = b_j+\beta_j.
\end{equation}
The resulting change in the component output is therefore
\begin{equation}
\left(H_jW_j'+\mathbf{1}{b_j'}^\top\right)
-
\left(H_jW_j+\mathbf{1}b_j^\top\right)
=
H_j\Delta W_j+\mathbf{1}\beta_j^\top.
\end{equation}
We choose the additive update so that this change matches $D_j$:
\begin{equation}
\min_{\Delta W_j,\beta_j}
\left\lVert
H_j\Delta W_j+\mathbf{1}\beta_j^{\top}-D_j
\right\rVert_F^2
+\lambda_j\left\lVert\Delta W_j\right\rVert_F^2.
\label{eq:direct_ridge}
\end{equation}
Centering the input and target banks removes the intercept from the weight solve. Let $\widetilde H_j=H_j-\mathbf{1}\mu_{H_j}^{\top}$ and $\widetilde D_j=D_j-\mathbf{1}\mu_{D_j}^{\top}$, where the means are column vectors. The closed-form ridge solution is
\begin{equation}
\Delta W_j=
\left(\widetilde H_j^\top\widetilde H_j+\lambda_j I\right)^{-1}
\widetilde H_j^\top\widetilde D_j,
\qquad
\beta_j=\mu_{D_j}-\Delta W_j^\top\mu_{H_j}.
\label{eq:direct_ridge_solution}
\end{equation}
Here, $H_j$ is collected from the target base model. Conditional on this fixed input bank, the additive update changes the selected linear component output by exactly
$H_j\Delta W_j+\mathbf{1}\beta_j^\top$.

\tit{Automatic Ridge Scaling.} Considering \cref{eq:direct_ridge}, rather than tuning $\lambda_j$, we set it from the calibration statistics using the empirical-Bayes ridge scale motivated by~\citet{srivastava2007comparison}:
\begin{equation}
\lambda_j=
\frac{\operatorname{tr}\!\left(
\widetilde H_j^\top \widetilde H_j
\right)}{n-1}.
\label{eq:empirical_bayes_inverse}
\end{equation}
This scales the regularization with the empirical magnitude of the component
inputs while shrinking directions weakly supported by the calibration data.
For a component with bias,
$\beta_j=\mu_{D_j}-\Delta W_j^\top\mu_{H_j}$.
For a bias-free component, we set $\beta_j=0$ and solve using
uncentered $H_j$ and $D_j$.

\tit{Returned task vector.} For $N$ selected blocks, we compile one update per block using \cref{eq:direct_ridge}. These regressions are independent: each $H_j$ is collected from the target base model, no fitted update is applied during optimization, and all updates are assembled only after every regression has been solved. This prevents the target of one regression from depending on the fitted parameters of another.

The resulting updates define a \textbf{sparse, compiled task vector} in the native target
parameterization, with non-zero entries only for the selected block components
introduced above. We denote this compiled task vector by $\tau_{\mathrm{t}}$,
which can be directly added to $\theta_{\mathrm{t}}^{\mathrm{base}}$.

\section{Experiments}\label{sec:experiments}

\tit{Models and tasks.}
We evaluate \method{} across vision and language models with heterogeneous architectures and scales. In vision, we transfer task updates between \texttt{DataComp-XL} OpenCLIP ViT-B/16 and ViT-L/14 checkpoints in both extension and reduction settings, evaluating performance on the Vision8 benchmark~\citep{ilharco2023editing}. In language, we consider \texttt{Qwen2.5} checkpoints~\citep{yang2024qwen2} at the 0.5B, 1.5B, and 3B parameter scales. We evaluate instruction-following transfer between the 0.5B and 1.5B models in both directions on IFEval~\citep{zhou2023instructionfollowing}, mathematical-reasoning transfer from \texttt{Qwen2.5-Math} 1.5B to the 0.5B and 3B base models on GSM8K~\citep{cobbe2021gsm8k}, and refusal-removal transfer between \texttt{Qwen2.5-Instruct} 0.5B and 1.5B, evaluated on HarmBench with ARC-Challenge as a capability control.

\begin{table*}[t]
\caption{\textbf{Task-vector transfer across transformer depth changes.}
We report ViT-B/16 $\rightarrow$ ViT-L/14 and ViT-L/14 $\rightarrow$ ViT-B/16 for $\mathcal{B}\in\{1,10\}$ calibration batches. Time is the median phase-total runtime per task, summing activation collection, transformation, and transport.}
\label{tab:main_transfer}
\centering
\setlength\tabcolsep{3.5pt}
\resizebox{\linewidth}{!}{%
\begin{tabular}{l c c c c c c c c c c c c c}
\toprule
& \textbf{No Grad.}
& $\mathcal{B}$
& ESAT
& SVHN
& GTSRB
& RESISC
& DTD
& Cars
& MNIST
& SUN
& \textbf{Avg ($\Delta$Acc)}
& \textbf{Time [s]}
& \textbf{Rel. Time} \\
\midrule

\grayrow
\multicolumn{14}{c}{
\textbf{ViT-B/16 $\rightarrow$ ViT-L/14 (Extension)}
} \\
\midrule

\multicolumn{14}{l}{\textit{Reference checkpoints}} \\

\quad Source base
& -- & -- & 55.84 & 62.57 & 55.20 & 69.17 & 57.45 & 88.89 & 76.49 & 70.01
& 66.95 & -- & -- \\

\quad Source fine-tuned
& -- & -- & 98.68 & 97.91 & 99.04 & 96.12 & 82.27 & 93.24 & 99.73 & 78.55
& 93.19 & -- & -- \\

\quad Target base
& -- & -- & 72.51 & 67.73 & 58.55 & 72.54 & 66.37 & 93.02 & 86.66 & 73.55
& 73.87 \deltazero & -- & -- \\

\midrule

\quad \theseus
& \greencmark
& 1
& 73.58 & 68.98 & 63.55 & 72.57 & 66.37 & 93.02 & 90.22 & 73.61
& 75.24 \deltapos{1.37}
& 1023
& 76.7 \\

\quad \bico
& \redxmark
& 1
& 72.10 & 70.57 & 61.67 & 72.54 & 66.37 & 93.02 & 93.73 & 73.55
& 75.44 \deltapos{1.58}
& 933
& 69.9 \\
\ourrow
\quad \method{}
& \greencmark
& 1
& 82.85 & 81.59 & 71.65 & 74.02 & 66.63 & 93.02 & 96.05 & 73.58
& 79.92 \deltapos{6.06}
& \textbf{13}
& \textbf{1} \\

\ourrow
\quad \methodgrad{}
& \redxmark
& 1
& 83.58 & 81.63 & 70.89 & 76.17 & 66.55 & 93.09 & 94.83 & 73.76
& \textbf{80.06 \deltapos{6.20}}
& \textbf{20}
& \textbf{1.5} \\

\midrule

\quad \theseus
& \greencmark
& 10
& 73.13 & 72.17 & 62.40 & 72.93 & 66.37 & 93.02 & 90.98 & 73.73
& 75.59 \deltapos{1.72}
& 1919
& 36.2 \\

\quad \bico
& \redxmark
& 10
& 74.44 & 71.55 & 67.90 & 73.35 & 66.37 & 93.02 & 96.11 & 73.63
& 77.05 \deltapos{3.18}
& 1998
& 37.7 \\

\ourrow
\quad \method{}
& \greencmark
& 10
& 87.50 & 90.93 & 84.17 & 79.01 & 69.48 & 93.02 & 98.47 & 73.77
& 84.54 \deltapos{10.68}
& \textbf{53}
& \textbf{1} \\

\ourrow
\quad \methodgrad{}
& \redxmark
& 10
& 90.78 & 90.22 & 87.60 & 84.96 & 72.40 & 93.62 & 98.24 & 74.19
& \textbf{86.50 \deltapos{12.63}}
& \textbf{80}
& \textbf{1.5} \\

\midrule

\grayrow
\multicolumn{14}{c}{
\textbf{ViT-L/14 $\rightarrow$ ViT-B/16 (Reduction)}
} \\
\midrule

\multicolumn{14}{l}{\textit{Reference checkpoints}} \\

\quad Source base
& -- & -- & 72.51 & 67.73 & 58.55 & 72.54 & 66.37 & 93.02 & 86.66 & 73.55
& 73.87 & -- & -- \\

\quad Source fine-tuned
& -- & -- & 99.14 & 98.30 & 99.20 & 96.98 & 85.70 & 94.93 & 99.76 & 82.62
& 94.58 & -- & -- \\

\quad Target base
& -- & -- & 55.84 & 62.57 & 55.20 & 69.17 & 57.45 & 88.89 & 76.49 & 70.01
& 66.95 \deltazero & -- & -- \\

\midrule

\quad \theseus
& \greencmark
& 1
& 55.80 & 63.49 & 55.24 & 68.43 & 57.45 & 88.89 & 76.49 & 70.01
& 66.98 \deltapos{0.02}
& 426
& 90.5 \\

\quad \bico
& \redxmark
& 1
& 56.75 & 62.60 & 55.89 & 68.92 & 57.27 & 88.89 & 79.43 & 69.99
& 67.47 \deltapos{0.52}
& 423
& 90.0 \\
\ourrow
\quad \method{}
& \greencmark
& 1
& 69.56 & 77.98 & 67.98 & 69.92 & 57.45 & 88.89 & 94.07 & 69.99
& 74.48 \deltapos{7.53}
& \textbf{4}
& \textbf{1} \\

\ourrow
\quad \methodgrad{}
& \redxmark
& 1
& 82.67 & 75.03 & 64.50 & 71.62 & 58.27 & 88.84 & 93.82 & 70.01
& \textbf{75.60 \deltapos{8.64}}
& \textbf{8}
& \textbf{1.8} \\

\midrule

\quad \theseus
& \greencmark
& 10
& 55.84 & 62.57 & 56.55 & 67.81 & 57.45 & 88.89 & 76.49 & 69.94
& 66.94 \deltaneg{0.01}
& 826
& 38.8 \\

\quad \bico
& \redxmark
& 10
& 59.38 & 62.54 & 56.43 & 70.34 & 57.51 & 88.89 & 79.39 & 70.25
& 68.09 \deltapos{1.14}
& 825
& 38.8 \\

\ourrow
\quad \method{}
& \greencmark
& 10
& 79.05 & 87.60 & 76.92 & 72.08 & 59.22 & 88.89 & 97.83 & 70.06
& 78.96 \deltapos{12.00}
& \textbf{21}
& \textbf{1} \\

\ourrow
\quad \methodgrad{}
& \redxmark
& 10
& 90.62 & 86.35 & 84.45 & 78.43 & 61.64 & 88.89 & 97.68 & 70.27
& \textbf{82.29 \deltapos{15.34}}
& \textbf{38}
& \textbf{1.8} \\

\bottomrule
\end{tabular}
}
\end{table*}

\tit{Baselines and protocol.}
We compare against the parameter-transport methods \theseus{} and \bico{} in both directions. Our default \method{} uses activation-based Procrustes alignment and therefore requires neither labels nor gradients; \methodgrad{} replaces this alignment signal with task-loss gradients for a supervision-matched comparison with \bico{}. We denote by $B$ the calibration budget, measured as the number of batches used to estimate the alignment and fit the target update, and use the same budget across methods. Task-vector and merge multipliers are selected on validation data, with test data reserved for final evaluation. The supplementary material reports additional results across calibration budgets, ridge choices, and resource measurements.

\subsection{Single-Task Vision Transfer}

\tit{Functional transfer succeeds in both directions.}
The central result in \cref{tab:main_transfer} is a simultaneous improvement in transfer accuracy and construction efficiency. Remarkably, with only $B=1$ calibration batch, \method{} already surpasses the best baseline using $B=10$ batches in both extension and reduction. Increasing the calibration budget further strengthens transfer, with \method{} improving over the target base by $10.68$ and $12.00$ points in extension and reduction, respectively. These gains are obtained while reducing fitting time by well over an order of magnitude: at the larger calibration budget, \method{} requires only $53$\,s for extension and $21$\,s for reduction, compared with roughly $32$\,min and $14$\,min for the parameter-transport baselines. The reported timing includes activation collection, transformation, and transport, and therefore covers the full measured fitting pipeline. As shown in \cref{tab:supp_phase_runtime}, most of the runtime gap arises from the transformation stage, while \cref{tab:supp_resource_scaling} shows that \method{} also requires substantially less peak host memory. The larger accuracy margin under reduction is particularly informative: \method{} cannot rely on additional target capacity or duplicated source computation, yet still recovers a substantial part of the source adaptation in a model with half as many blocks.

\begin{figure}[t]
    \centering
    \includegraphics[width=.9\linewidth]{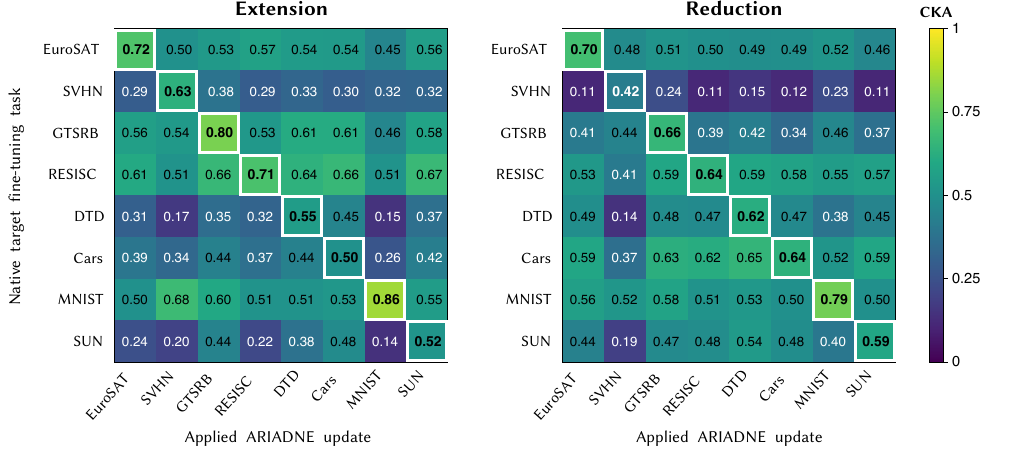}
    \vspace{-1em}
    \caption{\textbf{Task specificity of the transferred effect.} Debiased linear CKA between final-block CLS changes induced by native target fine-tuning (rows) and transported source fine-tuning (columns), for extension (left) and reduction (right). Updates use $\mathcal{B}=10$.}
    \label{fig:cross_task_cka}
\end{figure}

\tit{Alignment signal and calibration budget.}
The choice of alignment signal exposes a practical trade-off between supervision and transfer quality. At very small calibration budgets, gradient alignment provides a marginal advantage over activation alignment, but its benefit grows with additional labeled calibration data, as shown by the full calibration-budget sweep in \cref{tab:bank_alignment_per_task}. Thus, when labels are available and the extra cost of backward passes can be accommodated, \methodgrad{} provides the stronger option, whereas activation-based \method{} remains competitive while requiring neither labels nor gradients, making it a lighter default for unlabeled or resource-constrained calibration. More calibration generally improves both variants, though with diminishing returns. As detailed in \cref{tab:supp_d_only_extend,tab:supp_d_only_shrink}, increasing the calibration budget from one to ten batches yields substantial gains for activation-based \method{}, whereas further increases provide only modest improvements. Thus, most of the benefit is already captured with a modest calibration budget, making it a practical compromise between construction cost and the reliability of the resulting update.

\tit{Transferred updates produce task-specific representation changes.}
Accuracy alone does not show whether the transferred update preserves task-specific structure. We therefore measure each update's effect as the change in the final-block CLS representation relative to the target base. Using CKA, we compare these changes with those produced by native target fine-tuning on each of the eight tasks. In \cref{fig:cross_task_cka}, diagonal entries correspond to matching tasks, whereas off-diagonal entries compare different tasks. Each native fine-tuning effect is generally most similar to the update transferred from the same task, indicating that \method{} preserves task-specific representational changes rather than inducing a generic perturbation.

\begin{table*}[t]
\renewcommand{\arraystretch}{0.96}
\caption{\textbf{Component ablations.} \method{} ablations at $\mathcal{B}=10$.}
\label{tab:main_vision8_mechanism_ablation}
\centering
\setlength\tabcolsep{4pt}
\resizebox{\linewidth}{!}{%
\begin{tabular}{l c c c c c c c c c}
\toprule
\textbf{Variant}
& ESAT & SVHN & GTSRB & RESISC & DTD & Cars & MNIST & SUN
& \textbf{Avg ($\Delta$Acc)} \\
\midrule

\grayrow
\multicolumn{10}{c}{
\textbf{ViT-B/16 $\rightarrow$ ViT-L/14 (Extension)}
} \\
\midrule

\ourrow
\quad \method{}
& 87.98 & 91.02 & 83.26 & 78.87 & 68.68 & 93.02 & 98.50 & 73.73
& \textbf{84.38} \deltazero \\

\midrule

\multicolumn{10}{l}{\textit{Alignment and update components}} \\

\quad Random $Q_j$
& 72.35 & 67.73 & 59.43 & 72.56 & 66.08 & 93.13 & 86.66 & 73.55
& 73.93 \deltaneg{10.45} \\

\quad Ridge-regression $Q_j$
& 87.61 & 89.10 & 80.54 & 77.04 & 68.62 & 93.02 & 95.70 & 73.57
& 83.15 \deltaneg{1.23} \\

\quad Endpoint target
& 77.65 & 74.80 & 61.20 & 72.54 & 66.37 & 93.02 & 90.77 & 73.55
& 76.24 \deltaneg{8.14} \\

\midrule

\multicolumn{10}{l}{\textit{Block-pairing policy $\pi$}} \\

\quad Reversed pairing
& 82.96 & 71.45 & 64.73 & 73.76 & 67.38 & 92.99 & 91.88 & 73.53
& 77.34 \deltaneg{7.05} \\

\quad Pairing $+1$
& 87.01 & 90.31 & 82.80 & 78.87 & 68.50 & 93.02 & 98.36 & 73.72
& 84.07 \deltaneg{0.31} \\

\quad Pairing $-1$
& 83.62 & 88.53 & 77.56 & 75.89 & 67.55 & 93.02 & 97.46 & 73.60
& 82.15 \deltaneg{2.23} \\

\midrule

\grayrow
\multicolumn{10}{c}{
\textbf{ViT-L/14 $\rightarrow$ ViT-B/16 (Reduction)}
} \\
\midrule

\ourrow
\quad \method{}
& 79.30 & 88.76 & 75.92 & 72.36 & 60.22 & 88.89 & 98.23 & 70.01
& \textbf{79.21} \deltazero \\

\midrule

\multicolumn{10}{l}{\textit{Alignment and update components}} \\

\quad Random $Q_j$
& 55.84 & 62.57 & 55.02 & 68.62 & 57.45 & 88.89 & 76.49 & 69.37
& 66.78 \deltaneg{12.43} \\

\quad Ridge-regression $Q_j$
& 77.24 & 83.67 & 75.37 & 72.10 & 59.34 & 89.04 & 85.21 & 70.18
& 76.52 \deltaneg{2.69} \\

\quad Endpoint target
& 63.54 & 84.68 & 75.98 & 70.49 & 57.45 & 88.89 & 93.34 & 70.01
& 75.55 \deltaneg{3.66} \\

\midrule

\multicolumn{10}{l}{\textit{Block-pairing policy $\pi$}} \\

\quad Reversed pairing
& 65.80 & 73.71 & 58.79 & 69.52 & 58.04 & 88.89 & 86.88 & 70.01
& 71.46 \deltaneg{7.76} \\

\quad Pairing $+1$
& 79.34 & 88.12 & 75.46 & 72.54 & 59.75 & 88.88 & 98.20 & 70.04
& 79.04 \deltaneg{0.17} \\

\quad Pairing $-1$
& 74.32 & 88.71 & 75.18 & 71.06 & 60.11 & 88.89 & 98.18 & 70.02
& 78.31 \deltaneg{0.91} \\

\bottomrule
\end{tabular}
}
\end{table*}

\tit{What makes transfer work?}
\Cref{tab:main_vision8_mechanism_ablation} changes one part of the default configuration at a time. \emph{Random $Q_j$} replaces the learned Procrustes bridge with a random map, reducing averages essentially to the corresponding target-base performance. \emph{Ridge-regression $Q_j$} instead learns an unconstrained ridge map between the same base activations. It recovers most of the loss, but remains behind Procrustes, indicating that the orthogonality constraint is useful rather than merely convenient. \emph{Endpoint target} asks the target component to reproduce the aligned absolute source fine-tuned activation instead of only the fine-tuning-induced displacement (\cref{tab:bank_endpoint_target_per_task}). Its weaker performance supports transferring what fine-tuning \emph{changes}, while letting the target realize the displacement using the representation it already has, rather than directly reproducing source fine-tuned activations. Finally, we set the target-update regularizer to zero, but the resulting system is ill-conditioned: limited calibration and correlated features can leave the feature Gram matrix with small or zero eigenvalues, making the unregularized solve unstable. The empirical-Bayes scale avoids this failure without a sweep and remains a good heuristic relative to the fixed coefficients tested in either direction, as detailed in \cref{tab:supp_vision8_ridge_choice}.
The remaining rows alter only the block map $\pi$. \emph{Reversed pairing} flips the relative-depth order, matching shallow target blocks to deep source blocks and vice versa. \emph{Pairing} move every default match to the following or preceding source block, respectively. Both yield a drop in accuracy, with the reversed pairing being considerably worse. The correspondence prior can therefore be coarse and locally shifted, but it must respect the network's overall computational order.

\begin{table*}[t]
\vspace{-1.5em}
\def\arraystretch{1}
\caption{\textbf{\method{} for instruction tuning, mathematical reasoning and safety.}
We evaluate transfer on IFEval ($\mathcal{B}=100$), GSM8K ($\mathcal{B}=140$) and
safety HarmBench ($\mathcal{B}=100$).}
\label{tab:ariadne_instruction_math}
\centering
\resizebox{\linewidth}{!}{%
\begin{tabular}{l c c c c c c c c c c}
\toprule
& \multicolumn{3}{c}{\textbf{IFEval}}
& \multicolumn{3}{c}{\textbf{GSM8K}}
& \multicolumn{2}{c}{\textbf{HarmBench}}
& & \\
\textbf{Method}
& Inst. Strict 
& Prompt Strict
& \textbf{Avg.} 
& Strict 
& Flex. 
& \textbf{Avg.} 
& \textbf{Non-refusal}
& Cap. Check
& \textbf{Time [s]}
& \textbf{Rel. Time} \\
\midrule

\grayrow
\multicolumn{11}{c}{\textbf{Extension}} \\
\grayrow
& \multicolumn{3}{c}{\textbf{0.5B$\rightarrow$1.5B}}
& \multicolumn{3}{c}{\textbf{1.5B$\rightarrow$3B}}
& \multicolumn{2}{c}{\textbf{0.5B Inst$\rightarrow$1.5B Inst}}
& & \\
\midrule

Source base
& 23.41 & 15.62 & 19.52
& 61.40 & 62.02 & 61.70
& 54.17 & --
& -- & -- \\

Source fine-tuned
& 37.72 & 26.01 & 31.87
& 76.24 & 76.41 & 76.33
& 99.58 & --
& -- & -- \\

Target base
& 29.51 & 21.23 & 25.37
& 71.42 & 76.38 & 73.90
& 27.92 & 38.65
& -- & -- \\

\midrule

\quad \theseus
& 31.48 & 33.80 & 32.64~\deltapos{7.27}
& 72.24 & 73.61 & 72.93~\deltaneg{0.97}
& 27.50~\deltaneg{0.42} & 38.05
& 8702 & 6.65 \\

\quad \bico
& 36.41 & 27.93 & 32.17~\deltapos{6.80}
& 70.80 & 75.81 & 73.31~\deltaneg{0.59}
& 35.00~\deltapos{7.08} & \textbf{38.65}
& 9060 & 6.93 \\

\ourrow
\quad \method{}
& \textbf{42.59} & \textbf{44.44}
& \textbf{43.52}~\deltapos{18.15}
& \textbf{73.30} & \textbf{76.80}
& \textbf{75.05}~\deltapos{1.15}
& \textbf{67.92}~\deltapos{40.00} & 35.92
& \textbf{1308} & \textbf{1} \\

\midrule

\grayrow
\multicolumn{11}{c}{\textbf{Reduction}} \\
\grayrow
& \multicolumn{3}{c}{\textbf{1.5B$\rightarrow$0.5B}}
& \multicolumn{3}{c}{\textbf{1.5B$\rightarrow$0.5B}}
& \multicolumn{2}{c}{\textbf{1.5B Inst$\rightarrow$0.5B Inst} }
& & \\
\midrule

Source base
& 29.51 & 21.23 & 25.37
& 61.40 & 62.02 & 61.70
& 27.92 & --
& -- & -- \\

Source fine-tuned
& 40.56 & 47.09 & 43.83
& 76.24 & 76.41 & 76.33
& 93.33 & --
& -- & -- \\

Target base
& 23.41 & 15.62 & 19.52
& 32.83 & 33.61 & 33.22
& 54.17 & 28.58
& -- & -- \\

\midrule

\quad \theseus
& 24.07 & 26.85 & 25.46~\deltapos{5.94}
& 32.00 & 32.40 & 32.20~\deltaneg{1.02}
& 52.08~\deltaneg{2.09} & \textbf{27.99}
& 3746 & 7.14 \\

\quad \bico
& 25.00 & 28.70 & 26.85~\deltapos{7.33}
& \textbf{33.80} & 34.21
& 34.01~\deltapos{0.79}
& 54.58~\deltapos{0.41} & 27.98
& 3744 & 7.14 \\

\ourrow
\quad \method{}
& \textbf{30.09} & \textbf{31.02}
& \textbf{30.56}~\deltapos{11.04}
& 33.70 & \textbf{34.60}
& \textbf{34.15}~\deltapos{0.93}
& \textbf{67.50}~\deltapos{13.33} & 27.30
& \textbf{525} & \textbf{1} \\

\bottomrule
\end{tabular}
}
\end{table*}

\begin{table*}[t]
\def\arraystretch{.95}
\caption{\textbf{Merge-then-transport task-vector transfer.}
We first merge source task vectors and then
transport the resulting update to the target model using Vision8 calibration.
All methods use $\mathcal{B}=10$.}
\label{tab:merge_then_transport}
\centering
\setlength\tabcolsep{4pt}
\resizebox{\linewidth}{!}{%
\begin{tabular}{l c c c c c c c c c c}
\toprule
\textbf{Method}
& \textbf{Merge}
& EuroSAT
& SVHN
& GTSRB
& RESISC
& DTD
& Cars
& MNIST
& SUN
& \textbf{Avg ($\Delta$Acc)} \\
\midrule

\grayrow
\multicolumn{11}{c}{
\textbf{ViT-B/16 $\rightarrow$ ViT-L/14 (Extension)}
} \\
\midrule

\multicolumn{11}{l}{\textit{Reference checkpoint}} \\
\quad Target zero-shot
& --
& 72.51 & 67.73 & 58.55 & 72.54 & 66.37 & 93.02 & 86.66 & 73.55
& 73.87 \deltazero \\

\midrule

\quad \theseus
& TA
& 76.17 & 68.39 & 60.75 & 71.60 & 66.78 & 92.46 & 89.09 & 73.56
& 74.85 \deltapos{0.98} \\

\quad \bico
& TA
& 70.08 & 70.46 & 58.56 & 72.63 & 66.73 & 93.02 & 90.56 & 73.66
& 74.46 \deltapos{0.59} \\

\ourrow
\quad \method{}
& TA
& 75.27 & 73.14 & 63.46 & 72.45 & 64.95 & 92.75 & 90.01 & 73.43
& 75.68 \deltapos{1.82} \\

\midrule

\quad \theseus
& TSV
& 72.92 & 67.78 & 58.57 & 72.56 & 66.25 & 93.02 & 86.60 & 73.57
& 73.91 \deltapos{0.04} \\

\quad \bico
& TSV
& 71.36 & 69.41 & 58.85 & 72.77 & 66.67 & 93.02 & 89.73 & 73.56
& 74.42 \deltapos{0.55} \\

\ourrow
\quad \method{}
& TSV
& 82.67 & 78.04 & 69.86 & 73.44 & 66.02 & 92.83 & 96.66 & 73.34
& \textbf{79.11 \deltapos{5.24}} \\

\midrule

\grayrow
\multicolumn{11}{c}{
\textbf{ViT-L/14 $\rightarrow$ ViT-B/16 (Reduction)}
} \\
\midrule

\multicolumn{11}{l}{\textit{Reference checkpoint}} \\
\quad Target zero-shot
& --
& 55.84 & 62.57 & 55.20 & 69.17 & 57.45 & 88.89 & 76.49 & 70.01
& 66.95 \deltazero \\

\midrule

\quad \theseus
& TA
& 54.16 & 63.79 & 55.27 & 69.15 & 57.39 & 88.78 & 75.94 & 70.04
& 66.82 \deltaneg{0.13} \\

\quad \bico
& TA
& 54.65 & 62.76 & 55.21 & 68.99 & 57.51 & 88.99 & 76.62 & 70.11
& 66.86 \deltaneg{0.09} \\

\ourrow
\quad \method{}
& TA
& 54.77 & 66.71 & 59.09 & 69.31 & 57.45 & 88.70 & 79.53 & 70.09
& 68.21 \deltapos{1.25} \\

\midrule

\quad \theseus
& TSV
& 55.23 & 63.00 & 55.15 & 69.07 & 57.74 & 88.75 & 75.46 & 69.83
& 66.78 \deltaneg{0.17} \\

\quad \bico
& TSV
& 55.76 & 62.47 & 55.58 & 68.85 & 57.92 & 88.96 & 76.61 & 70.11
& 67.03 \deltapos{0.08} \\

\ourrow
\quad \method{}
& TSV
& 58.93 & 75.50 & 64.96 & 68.78 & 57.51 & 88.50 & 89.28 & 69.50
& \textbf{71.62 \deltapos{4.67}} \\

\bottomrule
\end{tabular}
}
\end{table*}

\subsection{Language-Model Transfer}

\Cref{tab:ariadne_instruction_math} reports transfer results, covering instruction following, mathematical reasoning, and safety alignment. In the \textit{Extension} setting, \method{} outperforms both baselines across all tasks. Notably, on IFEval, the transferred model even surpasses the source fine-tuned model, indicating that the transferred instruction-following behavior can combine with the stronger pre-trained capabilities of the target model. On the IFEval average, \method{} improves over the target base by $+18.15$ points, more than twice the gains achieved by other approaches. On GSM8K, \method{} is the only method that improves over the target base ($+1.15$), whereas both baselines slightly degrade performance. \method{} raises the HarmBench non-refusal rate from $27.92\%$ to $67.92\%$, while \theseus{} fails to transfer the behavior and \bico{} achieves only a modest gain ($+7.08$). This improvement comes at a small cost on the capability check ($-2.73$), suggesting that general capabilities are largely preserved.
In the \textit{Reduction} setting, the results are less pronounced, but \method{} again achieves the best performance across all settings. While the gain is more noticeable on IFEval, on GSM8K all methods remain within approximately one point of the target base. We attribute this to the smaller target model's limited capacity: its base accuracy is substantially lower than that of the source fine-tuned model, leaving limited room for the transferred reasoning skills to manifest. On HarmBench, \method{} is the only method that successfully transfers the abliteration to the smaller model, whereas \theseus{} and \bico{} leave the target's behavior essentially unchanged.

\subsection{Multi-Task Transfer}

We finally ask whether one fit can transfer a source checkpoint that already combines all eight task vectors. We merge in source space with Task Arithmetic (TA)~\citep{ilharco2023editing} or Task Singular Vectors (TSV)~\citep{gargiulo2025task}, apply the merged vector to the source base, and fit \method{} to its joint functional effect. This \emph{merge-then-transfer} strategy requires one fit, whereas transferring each vector before merging requires eight.

\Cref{tab:merge_then_transport} shows that \method{} is the only approach that improves the target average for both merge rules and both directions. The choice of source merge is consequential: TSV raises \method{} from $75.68\%$ to $79.11\%$ in extension and from $68.21\%$ to $71.62\%$ in reduction. In contrast, changing from TA to TSV does not consistently help the transport baselines. Direct functional transfer can preserve a good merged effect, but it cannot undo interference already introduced by a poorer source merge. The comparison with transfer-then-merge in \cref{tab:transport_then_merge} makes the efficiency trade-off explicit. Fitting vectors separately is better for TA by $2.73$ points in extension and $4.15$ points in reduction, and better for TSV reduction by $1.01$ points. With TSV extension, however, merging first is $0.17$ points better despite using one eighth as many fits. Merge-then-transfer is therefore not uniformly accuracy-equivalent but it is most attractive when the source-side merger already produces a coherent joint effect. Moreover, some datasets remain at or slightly below the target base even when the average improves, showing that multi-task interference is reduced rather than eliminated.

\section{Conclusions and Limitations}
We introduced \method{}, a training-free method for transferring fine-tuning effects across Transformers with different depths and widths. By aligning source activation changes and fitting target-native updates in closed form, \method{} avoids parameter transport while improving vision, language, and multi-task transfer at substantially lower construction cost. Our results support local functional change as a practical object for cross-architecture transfer.

Our evaluation covers related vision and language backbones within two model families, so it does not establish transfer across arbitrary architectures or modalities. The method also requires representative paired calibration inputs and a block correspondences. These limitations motivate more adaptive correspondence and update selection, as well as broader cross-family evaluation.

\subsection*{AI use statement}

In this work, we used generative AI tools to assist with the grammar, wording, and structure of the manuscript, as well as with writing and debugging code. All AI-assisted text was reviewed and revised by the authors, and all AI-assisted code was inspected and validated through the relevant experiments and checks. Generative AI was not used to produce experimental data or to make final scientific decisions. We take responsibility for the complete content of this work, including all text, claims, code, and results produced with the aid of generative AI.

\bibliography{bibliography}

@String(CVPR= {IEEE Conf. Comput. Vis. Pattern Recog.})

@String(ICLR = {Int. Conf. Learn. Represent.})

@String(CVPR  = {CVPR})

@String(ICLR  = {ICLR})

@string{anips = {Advances in Neural Information Processing Systems}}

@string{iclr = {International Conference on Learning Representations}}

@string{neuripsw = {Neural Information Processing Systems Workshops}}

@string{icml = {International Conference on Machine Learning}}

@string{cvpr = {Proceedings of the IEEE conference on Computer Vision and Pattern Recognition}}

@string{emnlp = {Empirical Methods in Natural Language Processing}}

@string{iccvw = {IEEE International Conference on Computer Vision Workshops}}

@inproceedings{kangaslahti2026boomerang,
  title={Boomerang distillation enables zero-shot model size interpolation},
  author={Kangaslahti, Sara and Nayak, Nihal V and Geuter, Jonathan and Fumero, Marco and Locatello, Francesco and Alvarez-Melis, David},
  booktitle=iclr,
  volume={2026},
  pages={73351--73389},
  year={2026}
}

@article{cobbe2021gsm8k,
  title={Training Verifiers to Solve Math Word Problems},
  author={Cobbe, Karl and Kosaraju, Vineet and Bavarian, Mohammad and Chen, Mark and Jun, Heewoo and Kaiser, Lukasz and Plappert, Matthias and Tworek, Jerry and Hilton, Jacob and Nakano, Reiichiro and Hesse, Christopher and Schulman, John},
  journal={arXiv preprint arXiv:2110.14168},
  year={2021}
}

@article{yang2024qwen2,
  title={Qwen2. 5-math technical report: Toward mathematical expert model via self-improvement},
  author={Yang, An and Zhang, Beichen and Hui, Binyuan and Gao, Bofei and Yu, Bowen and Li, Chengpeng and Liu, Dayiheng and Tu, Jianhong and Zhou, Jingren and Lin, Junyang and others},
  journal={arXiv preprint arXiv:2409.12122},
  year={2024}
}

@inproceedings{ilharco2023editing,
  author       = {Gabriel Ilharco and
                  Marco T{\'{u}}lio Ribeiro and
                  Mitchell Wortsman and
                  Ludwig Schmidt and
                  Hannaneh Hajishirzi and
                  Ali Farhadi},
  title        = {Editing models with task arithmetic},
  booktitle    = ICLR,
  year         = {2023}
}

@inproceedings{cars,
  title={{3D Object representations for fine-grained categorization}},
  author={Krause, Jonathan and Stark, Michael and Deng, Jia and Fei-Fei, Li},
  booktitle=iccvw,
  year={2013}
}

@inproceedings{svhn,
  title={Reading digits in natural images with unsupervised feature learning},
  author={Netzer, Yuval and Wang, Tao and Coates, Adam and Bissacco, Alessandro and Wu, Bo and Ng, Andrew Y},
  booktitle=neuripsw,
  year={2011}
}

@inproceedings{gtsrb,
  title={The German traffic sign recognition benchmark: a multi-class classification competition},
  author={Stallkamp, Johannes and Schlipsing, Marc and Salmen, Jan and Igel, Christian},
  booktitle={IJCNN},
  year={2011},
}

@inproceedings{radford2021learning,
    Author = {Alec Radford and Jong Wook Kim and Chris Hallacy and Aditya Ramesh and Gabriel Goh and Sandhini Agarwal and Girish Sastry and Amanda Askell and Pamela Mishkin and Jack Clark and Gretchen Krueger and Ilya Sutskever},
    Title = {Learning Transferable Visual Models From Natural Language Supervision},
    Year = {2021},
    booktitle=icml,
}

@inproceedings{ainsworth2023git,
  author       = {Samuel K. Ainsworth and
                  Jonathan Hayase and
                  Siddhartha S. Srinivasa},
  title        = {Git Re-Basin: Merging Models modulo Permutation Symmetries},
  booktitle    = iclr,
  year         = {2023},
 }

@misc{zhou2023instructionfollowing,
      title={Instruction-Following Evaluation for Large Language Models}, 
      author={Jeffrey Zhou and Tianjian Lu and Swaroop Mishra and Siddhartha Brahma and Sujoy Basu and Yi Luan and Denny Zhou and Le Hou},
      year={2023},
      eprint={2311.07911},
      archivePrefix={arXiv},
      primaryClass={cs.CL},
      url={https://arxiv.org/abs/2311.07911}, 
}

@article{yang2025qwen3,
  title={Qwen3 technical report},
  author={Yang, An and Li, Anfeng and Yang, Baosong and Zhang, Beichen and Hui, Binyuan and Zheng, Bo and Yu, Bowen and Gao, Chang and Huang, Chengen and Lv, Chenxu and others},
  journal={arXiv preprint arXiv:2505.09388},
  year={2025}
}

@inproceedings{jin2023regmean,
  author    = {Xisen Jin and Xiang Ren and Daniel Preotiuc{-}Pietro and Pengxiang Cheng},
  title     = {Dataless Knowledge Fusion by Merging Weights of Language Models},
  booktitle = iclr,
  year      = {2023},
}

@article{rinaldi2025update,
  title   = {Update Your Transformer to the Latest Release: Re-Basin of Task Vectors},
  author  = {Filippo Rinaldi and Giacomo Capitani and Lorenzo Bonicelli and Donato Crisostomi and Federico Bolelli and Elisa Ficarra and Emanuele Rodolà and Simone Calderara and Angelo Porrello},
  year    = {2025},
  journal = icml,
}

@inproceedings{nasery2025pleas,
  title={PLeaS-Merging Models with Permutations and Least Squares},
  author={Nasery, Anshul and Hayase, Jonathan and Koh, Pang Wei and Oh, Sewoong},
  booktitle=cvpr,
  year={2025}
}

@inproceedings{chen2015net2net,
  title={Net2net: Accelerating learning via knowledge transfer},
  author={Chen, Tianqi and Goodfellow, Ian and Shlens, Jonathon},
  booktitle=iclr,
  year={2015}
}

@inproceedings{wei2016network,
  title={Network morphism},
  author={Wei, Tao and Wang, Changhu and Rui, Yong and Chen, Chang Wen},
  booktitle=icml,
  year={2016},
}

@article{hinton2015distilling,
  title   = {Distilling the Knowledge in a Neural Network},
  author  = {Geoffrey Hinton and Oriol Vinyals and Jeff Dean},
  year    = {2015},
  journal = neuripsw
}

@inproceedings{rinaldi2026gradient,
title={Gradient-Sign Masking for Task Vector Transport Across Pre-Trained Models},
author={Filippo Rinaldi and Aniello Panariello and Giacomo Salici and Fengyuang Liu and Marco Ciccone and Angelo Porrello and Simone Calderara},
booktitle=iclr,
year={2026}
}

@inproceedings{williams2021generalized,
  title   = {Generalized shape metrics on neural representations},
  author  = {Williams, Alex H and Kunz, Erin and Kornblith, Simon and Linderman, Scott},
  booktitle=anips,
  year    = {2021}
}

@inproceedings{
    moschella2023relative,
    title={Relative representations enable zero-shot latent space communication},
    author={Luca Moschella and Valentino Maiorca and Marco Fumero and Antonio Norelli and Francesco Locatello and Emanuele Rodol{\`a}},
    booktitle=iclr,
    year={2023},
}

@inproceedings{
maiorca2023latent,
title={Latent Space Translation via Semantic Alignment},
author={Valentino Maiorca and Luca Moschella and Antonio Norelli and Marco Fumero and Francesco Locatello and Emanuele Rodol{\`a}},
booktitle=anips,
year={2023},
}

@article{gesmundo2023composable,
  title={Composable function-preserving expansions for transformer architectures},
  author={Gesmundo, Andrea and Maile, Kaitlin},
  journal={arXiv preprint arXiv:2308.06103},
  year={2023}
}

@inproceedings{ding2023network,
  title={Network expansion for practical training acceleration},
  author={Ding, Ning and Tang, Yehui and Han, Kai and Xu, Chao and Wang, Yunhe},
  booktitle=cvpr,
  year={2023},
}

@inproceedings{chen2022bert2bert,
  title={bert2bert: Towards reusable pretrained language models},
  author={Chen, Cheng and Yin, Yichun and Shang, Lifeng and Jiang, Xin and Qin, Yujia and Wang, Fengyu and Wang, Zhi and Chen, Xiao and Liu, Zhiyuan and Liu, Qun},
  booktitle={Proceedings of the Association for Computational Linguistics},
  year={2022}
}

@inproceedings{fan2020Reducing,
title={Reducing Transformer Depth on Demand with Structured Dropout},
author={Angela Fan and Edouard Grave and Armand Joulin},
booktitle=iclr,
year={2020},
}

@article{sajjad2023effect,
  title={On the effect of dropping layers of pre-trained transformer models},
  author={Sajjad, Hassan and Dalvi, Fahim and Durrani, Nadir and Nakov, Preslav},
  journal={Computer Speech \& Language},
  year={2023},
}

@inproceedings{xu2020bert,
  title={Bert-of-theseus: Compressing bert by progressive module replacing},
  author={Xu, Canwen and Zhou, Wangchunshu and Ge, Tao and Wei, Furu and Zhou, Ming},
  booktitle=emnlp,
  year={2020}
}

@article{son2026bico,
  title={Bilinear Coordinate Alignment for Training-Free Task-Vector Transfer},
  author={Son, Jungyong and Jung, Jinwook and Park, Minhee and Baik, Sungyong},
  journal={arXiv preprint arXiv:2605.28444},
  year={2026}
}

@article{rinaldi2026transporting,
  title   = {Transporting Task Vectors across Different Architectures without Training},
  author  = {Filippo Rinaldi and Aniello Panariello and Giacomo Salici and Angelo Porrello and Simone Calderara},
  year    = {2026},
  journal = icml
}

@article{srivastava2007comparison,
  title={Comparison of discrimination methods for high dimensional data},
  author={Srivastava, Muni S and Kubokawa, Tatsuya},
  journal={Journal of the Japan Statistical Society},
  year={2007},
}

@inproceedings{gargiulo2025task,
  title={Task singular vectors: Reducing task interference in model merging},
  author={Gargiulo, Antonio Andrea and Crisostomi, Donato and Bucarelli, Maria Sofia and Scardapane, Simone and Silvestri, Fabrizio and Rodola, Emanuele},
  booktitle=cvpr,
  year={2025},
}

@article{bansal2021revisiting,
  title={Revisiting model stitching to compare neural representations},
  author={Bansal, Yamini and Nakkiran, Preetum and Barak, Boaz},
  journal={Advances in neural information processing systems},
  year={2021}
}

@inproceedings{jiao2020tinybert,
  title={Tinybert: Distilling bert for natural language understanding},
  author={Jiao, Xiaoqi and Yin, Yichun and Shang, Lifeng and Jiang, Xin and Chen, Xiao and Li, Linlin and Wang, Fang and Liu, Qun},
  booktitle={Findings of the association for computational linguistics: EMNLP 2020},
  pages={4163--4174},
  year={2020}
}
\bibliographystyle{iclr2027_conference}

\clearpage
\appendix
\section{Appendix}
\subsection{Calibration Budget and Stability}

\Cref{tab:supp_d_only_extend,tab:supp_d_only_shrink} expand the main comparison across five calibration budgets and three calibration runs. Accuracy improves rapidly at small budgets: moving from one to five batches raises the mean from 79.92\% to 83.54\% in extension and from 74.48\% to 77.97\% in reduction. The subsequent gains are smaller, reaching 85.68\% and 80.20\% at 50 batches. Additional observations therefore refine the fitted effect rather than changing the qualitative outcome. They also improve stability in the harder reduction setting: its average standard deviation falls from 0.98 at one batch to 0.24 at 50, with the especially variable EuroSAT and GTSRB fits becoming more consistent. The benefit is task-dependent, however. EuroSAT, SVHN, GTSRB, and MNIST continue to improve with calibration, whereas Cars and SUN397 remain near their target-base values. More calibration reduces estimation error but cannot make every source effect useful to the target.

\subsection{Transport Then Merge}

\Cref{tab:transport_then_merge} reports the more expensive alternative to the merge-then-transfer strategy in the main paper: each of the eight task vectors is first transferred independently, after which the target vectors are merged. \method{} improves on the target base for both merge rules and directions, gaining 4.54--5.08 points in extension and 5.40--5.68 points in reduction. Once tasks have been transferred separately, TA and TSV differ by at most 0.53 points, much less than when the source vectors are merged before transfer. This suggests that separate functional fits reduce the source-side interference that makes the merge rule consequential in \cref{tab:merge_then_transport}. The improvement comes at the cost of eight fits rather than one, and it is not uniform across datasets: gains on EuroSAT, SVHN, GTSRB, and MNIST coexist with small losses on some of DTD, Cars, RESISC45, and SUN397. The average therefore captures useful multi-task transfer, not Pareto improvement on every task.

\subsection{Activation and Gradient Alignment}

\Cref{tab:bank_alignment_per_task} separates the alignment signal from the direct-regression objective. With one calibration batch, activation- and gradient-derived alignments are essentially tied in both directions. Their gap grows with the budget: at 20 batches, gradient alignment is 2.41 points better in extension and 4.20 points better in reduction. Gradients are especially helpful for EuroSAT, GTSRB, RESISC45, and DTD, but do not dominate on every task; activation alignment remains competitive on SVHN and MNIST at several budgets. The gradient variant is thus useful when labels and backward passes are available, while activation alignment retains the practical advantage of label-free calibration.

\begin{table*}[t]
\caption{\textbf{Single-task transfer across calibration budgets: Extension.}
We report Vision8 test accuracy (\%) for \method{} across calibration budgets
$\mathcal{B}\in\{1,5,10,20,50\}$.
Results are mean $\pm$ standard deviation over three calibration runs.
$\Delta$Acc denotes the improvement over the target zero-shot checkpoint.}
\label{tab:supp_d_only_extend}
\centering
\setlength\tabcolsep{4pt}
\resizebox{\linewidth}{!}{%
\begin{tabular}{l c c c c c c c c c c}
\toprule
\textbf{Method}
& $\mathcal{B}$
& EuroSAT
& SVHN
& GTSRB
& RESISC
& DTD
& Cars
& MNIST
& SUN
& \textbf{Avg ($\Delta$Acc)} \\
\midrule

\multicolumn{11}{l}{\textit{Reference checkpoint}} \\
\quad Target zero-shot
& --
& 72.51
& 67.73
& 58.55
& 72.54
& 66.37
& 93.02
& 86.66
& 73.55
& 73.87 \deltazero \\

\midrule

\ourrow
\quad \method{}
& 1
& 82.85{\scriptsize$\pm$0.82}
& 81.59{\scriptsize$\pm$2.02}
& 71.65{\scriptsize$\pm$1.47}
& 74.02{\scriptsize$\pm$0.41}
& 66.63{\scriptsize$\pm$0.24}
& 93.02{\scriptsize$\pm$0.00}
& 96.05{\scriptsize$\pm$0.54}
& 73.58{\scriptsize$\pm$0.18}
& \textbf{79.92{\scriptsize$\pm$0.32} \deltapos{6.06}} \\

\midrule

\ourrow
\quad \method{}
& 5
& 86.15{\scriptsize$\pm$1.84}
& 89.84{\scriptsize$\pm$0.12}
& 81.64{\scriptsize$\pm$1.45}
& 77.56{\scriptsize$\pm$0.62}
& 68.40{\scriptsize$\pm$0.44}
& 93.02{\scriptsize$\pm$0.00}
& 98.03{\scriptsize$\pm$0.12}
& 73.67{\scriptsize$\pm$0.10}
& \textbf{83.54{\scriptsize$\pm$0.11} \deltapos{9.67}} \\

\midrule

\ourrow
\quad \method{}
& 10
& 87.50{\scriptsize$\pm$0.44}
& 90.93{\scriptsize$\pm$0.24}
& 84.17{\scriptsize$\pm$0.96}
& 79.01{\scriptsize$\pm$0.20}
& 69.48{\scriptsize$\pm$0.70}
& 93.02{\scriptsize$\pm$0.00}
& 98.47{\scriptsize$\pm$0.08}
& 73.77{\scriptsize$\pm$0.04}
& \textbf{84.54{\scriptsize$\pm$0.19} \deltapos{10.68}} \\

\midrule

\ourrow
\quad \method{}
& 20
& 87.76{\scriptsize$\pm$1.22}
& 91.82{\scriptsize$\pm$0.26}
& 86.71{\scriptsize$\pm$0.49}
& 79.13{\scriptsize$\pm$0.06}
& 70.57{\scriptsize$\pm$0.53}
& 93.02{\scriptsize$\pm$0.00}
& 98.66{\scriptsize$\pm$0.09}
& 73.70{\scriptsize$\pm$0.03}
& \textbf{85.17{\scriptsize$\pm$0.07} \deltapos{11.31}} \\

\midrule

\ourrow
\quad \method{}
& 50
& 89.38{\scriptsize$\pm$1.28}
& 92.61{\scriptsize$\pm$0.41}
& 87.25{\scriptsize$\pm$1.13}
& 79.79{\scriptsize$\pm$0.07}
& 70.78{\scriptsize$\pm$1.29}
& 93.02{\scriptsize$\pm$0.00}
& 98.84{\scriptsize$\pm$0.07}
& 73.80{\scriptsize$\pm$0.04}
& \textbf{85.68{\scriptsize$\pm$0.19} \deltapos{11.82}} \\

\bottomrule
\end{tabular}
}
\end{table*}

\begin{table*}[t]
\caption{\textbf{Single-task transfer across calibration budgets: Reduction.}
We report Vision8 test accuracy (\%) for \method{} across calibration budgets
$\mathcal{B}\in\{1,5,10,20,50\}$.
Results are mean $\pm$ standard deviation over three calibration runs.
$\Delta$Acc denotes the improvement over the target zero-shot checkpoint.}
\label{tab:supp_d_only_shrink}
\centering
\setlength\tabcolsep{4pt}
\resizebox{\linewidth}{!}{%
\begin{tabular}{l c c c c c c c c c c}
\toprule
\textbf{Method}
& $\mathcal{B}$
& EuroSAT
& SVHN
& GTSRB
& RESISC
& DTD
& Cars
& MNIST
& SUN
& \textbf{Avg ($\Delta$Acc)} \\
\midrule

\multicolumn{11}{l}{\textit{Reference checkpoint}} \\
\quad Target zero-shot
& --
& 55.84
& 62.57
& 55.20
& 69.17
& 57.45
& 88.89
& 76.49
& 70.01
& 66.95 \deltazero \\

\midrule

\ourrow
\quad \method{}
& 1
& 69.56{\scriptsize$\pm$5.72}
& 77.98{\scriptsize$\pm$1.38}
& 67.98{\scriptsize$\pm$0.67}
& 69.92{\scriptsize$\pm$0.62}
& 57.45{\scriptsize$\pm$0.00}
& 88.89{\scriptsize$\pm$0.00}
& 94.07{\scriptsize$\pm$0.91}
& 69.99{\scriptsize$\pm$0.11}
& \textbf{74.48{\scriptsize$\pm$0.98} \deltapos{7.53}} \\

\midrule

\ourrow
\quad \method{}
& 5
& 76.47{\scriptsize$\pm$0.58}
& 85.67{\scriptsize$\pm$1.68}
& 74.47{\scriptsize$\pm$1.85}
& 72.00{\scriptsize$\pm$0.50}
& 58.59{\scriptsize$\pm$1.02}
& 88.93{\scriptsize$\pm$0.06}
& 97.64{\scriptsize$\pm$0.14}
& 70.01{\scriptsize$\pm$0.01}
& \textbf{77.97{\scriptsize$\pm$0.55} \deltapos{11.02}} \\

\midrule

\ourrow
\quad \method{}
& 10
& 79.05{\scriptsize$\pm$0.77}
& 87.60{\scriptsize$\pm$1.41}
& 76.92{\scriptsize$\pm$3.01}
& 72.08{\scriptsize$\pm$0.50}
& 59.22{\scriptsize$\pm$1.54}
& 88.89{\scriptsize$\pm$0.00}
& 97.83{\scriptsize$\pm$0.35}
& 70.06{\scriptsize$\pm$0.07}
& \textbf{78.96{\scriptsize$\pm$0.73} \deltapos{12.00}} \\

\midrule

\ourrow
\quad \method{}
& 20
& 80.70{\scriptsize$\pm$1.55}
& 88.47{\scriptsize$\pm$0.78}
& 77.95{\scriptsize$\pm$2.71}
& 72.56{\scriptsize$\pm$0.26}
& 58.55{\scriptsize$\pm$1.91}
& 88.89{\scriptsize$\pm$0.00}
& 98.10{\scriptsize$\pm$0.09}
& 70.09{\scriptsize$\pm$0.03}
& \textbf{79.41{\scriptsize$\pm$0.41} \deltapos{12.46}} \\

\midrule

\ourrow
\quad \method{}
& 50
& 81.44{\scriptsize$\pm$0.53}
& 88.73{\scriptsize$\pm$0.50}
& 79.56{\scriptsize$\pm$0.73}
& 73.83{\scriptsize$\pm$0.33}
& 60.84{\scriptsize$\pm$0.54}
& 88.89{\scriptsize$\pm$0.00}
& 98.25{\scriptsize$\pm$0.03}
& 70.05{\scriptsize$\pm$0.06}
& \textbf{80.20{\scriptsize$\pm$0.24} \deltapos{13.24}} \\

\bottomrule
\end{tabular}
}
\end{table*}

\begin{table*}[t]
\caption{\textbf{Transport-then-merge task-vector transfer.}
We first transport individual task vectors to the target model using
\method{}, \theseus, or \bico and then merge them using Task Arithmetic (TA)
or TSV. All methods use $\mathcal{B}=10$ calibration batches and a single
shared validation-selected merge coefficient.
$\Delta$Acc denotes the improvement over the target zero-shot checkpoint.}
\label{tab:transport_then_merge}
\centering
\setlength\tabcolsep{4pt}
\resizebox{\linewidth}{!}{%
\begin{tabular}{l c c c c c c c c c c}
\toprule
\textbf{Method}
& \textbf{Merge}
& EuroSAT
& SVHN
& GTSRB
& RESISC
& DTD
& Cars
& MNIST
& SUN
& \textbf{Avg ($\Delta$Acc)} \\
\midrule

\grayrow
\multicolumn{11}{c}{
\textbf{ViT-B/16 $\rightarrow$ ViT-L/14 (Extension)}
} \\
\midrule

\multicolumn{11}{l}{\textit{Reference checkpoint}} \\
\quad Target zero-shot
& --
& 72.51 & 67.73 & 58.55 & 72.54 & 66.37 & 93.02 & 86.66 & 73.55
& 73.87 \deltazero \\

\midrule

\quad \theseus
& TA
& 73.95 & 70.19 & 60.51 & 73.00 & 66.31 & 92.98 & 88.56 & 73.73
& 74.90 \deltapos{1.03} \\

\quad \bico
& TA
& 71.60 & 71.36 & 63.83 & 72.73 & 68.09 & 92.61 & 94.66 & 73.60
& 76.06 \deltapos{2.19} \\

\ourrow
\quad \method{}
& TA
& 79.59 & 78.91 & 67.92 & 73.67 & 65.66 & 92.44 & 95.83 & 73.26
& \textbf{78.41 \deltapos{4.54}} \\

\midrule

\quad \theseus
& TSV
& 72.59 & 67.76 & 58.56 & 72.54 & 66.37 & 93.02 & 86.69 & 73.54
& 73.88 \deltapos{0.02} \\

\quad \bico
& TSV
& 74.69 & 72.71 & 67.63 & 71.31 & 66.43 & 92.69 & 96.80 & 73.33
& 76.95 \deltapos{3.08} \\

\ourrow
\quad \method{}
& TSV
& 84.86 & 80.10 & 67.89 & 71.69 & 66.31 & 91.29 & 96.98 & 72.42
& \textbf{78.94 \deltapos{5.08}} \\

\midrule

\grayrow
\multicolumn{11}{c}{
\textbf{ViT-L/14 $\rightarrow$ ViT-B/16 (Reduction)}
} \\
\midrule

\multicolumn{11}{l}{\textit{Reference checkpoint}} \\
\quad Target zero-shot
& --
& 55.84 & 62.57 & 55.20 & 69.17 & 57.45 & 88.89 & 76.49 & 70.01
& 66.95 \deltazero \\

\midrule

\quad \theseus
& TA
& 55.84 & 62.58 & 55.25 & 69.17 & 57.57 & 88.88 & 76.38 & 70.01
& 66.96 \deltapos{0.01} \\

\quad \bico
& TA
& 56.05 & 61.81 & 55.39 & 69.58 & 57.45 & 88.84 & 77.81 & 70.11
& 67.13 \deltapos{0.18} \\

\ourrow
\quad \method{}
& TA
& 62.88 & 74.80 & 67.01 & 69.12 & 57.86 & 87.76 & 90.61 & 68.82
& \textbf{72.36 \deltapos{5.40}} \\

\midrule

\quad \theseus
& TSV
& 56.09 & 62.66 & 55.25 & 69.12 & 57.39 & 88.89 & 76.31 & 70.01
& 66.96 \deltapos{0.01} \\

\quad \bico
& TSV
& 56.05 & 62.17 & 55.29 & 69.66 & 57.80 & 88.89 & 77.72 & 70.00
& 67.20 \deltapos{0.24} \\

\ourrow
\quad \method{}
& TSV
& 65.14 & 76.18 & 65.80 & 67.23 & 59.10 & 84.87 & 94.17 & 68.56
& \textbf{72.63 \deltapos{5.68}} \\

\bottomrule
\end{tabular}
}
\end{table*}

\begin{table*}[t]
\caption{\textbf{Activation- versus gradient-based alignment.}
We compare activation-derived $Q_{\mathrm{act}}$ and gradient-derived
$Q_{\mathrm{grad}}$ within \method{} across calibration budgets
$\mathcal{B}\in\{1,2,5,10,20\}$.
Entries are Vision8 test accuracy (\%).
$\Delta$Acc denotes the improvement over the target zero-shot checkpoint.}
\label{tab:bank_alignment_per_task}
\centering
\setlength\tabcolsep{4pt}
\resizebox{\linewidth}{!}{%
\begin{tabular}{l c c c c c c c c c c}
\toprule
& $\mathcal{B}$
& EuroSAT
& SVHN
& GTSRB
& RESISC
& DTD
& Cars
& MNIST
& SUN
& \textbf{Avg ($\Delta$Acc)} \\
\midrule

\grayrow
\multicolumn{11}{c}{
\textbf{ViT-B/16 $\rightarrow$ ViT-L/14 (Extension)}
} \\
\midrule

\multicolumn{11}{l}{\textit{Reference checkpoints}} \\

\quad Source zero-shot
& -- & 55.84 & 62.57 & 55.20 & 69.17 & 57.45 & 88.89 & 76.49 & 70.01
& 66.95 \\

\quad Source fine-tuned
& -- & 98.68 & 97.91 & 99.04 & 96.12 & 82.27 & 93.24 & 99.73 & 78.55
& 93.19 \\

\quad Target zero-shot
& -- & 72.51 & 67.73 & 58.55 & 72.54 & 66.37 & 93.02 & 86.66 & 73.55
& 73.87 \deltazero \\

\midrule

\quad $Q_{\mathrm{act}}$
& 1 & 82.22 & 83.41 & 70.00 & 73.97 & 66.37 & 93.02 & 95.54 & 73.71
& 79.78 \deltapos{5.91} \\

\ourrow
\quad $Q_{\mathrm{grad}}$
& 1 & 83.58 & 81.63 & 70.89 & 76.17 & 66.55 & 93.09 & 94.83 & 73.76
& \textbf{80.06 \deltapos{6.20}} \\

\midrule

\quad $Q_{\mathrm{act}}$
& 2 & 86.46 & 87.86 & 76.29 & 76.38 & 68.62 & 93.02 & 96.47 & 73.53
& 82.33 \deltapos{8.46} \\

\ourrow
\quad $Q_{\mathrm{grad}}$
& 2 & 89.42 & 84.30 & 81.84 & 78.31 & 67.43 & 93.24 & 96.24 & 73.89
& \textbf{83.09 \deltapos{9.22}} \\

\midrule

\quad $Q_{\mathrm{act}}$
& 5 & 88.23 & 89.73 & 79.99 & 78.15 & 68.74 & 93.02 & 97.91 & 73.55
& 83.67 \deltapos{9.80} \\

\ourrow
\quad $Q_{\mathrm{grad}}$
& 5 & 89.59 & 87.42 & 81.83 & 82.40 & 72.70 & 93.37 & 97.72 & 74.12
& \textbf{84.89 \deltapos{11.02}} \\

\midrule

\quad $Q_{\mathrm{act}}$
& 10 & 87.98 & 91.02 & 83.26 & 78.87 & 68.68 & 93.02 & 98.50 & 73.73
& 84.38 \deltapos{10.51} \\

\ourrow
\quad $Q_{\mathrm{grad}}$
& 10 & 90.78 & 90.22 & 87.60 & 84.96 & 72.40 & 93.62 & 98.24 & 74.19
& \textbf{86.50 \deltapos{12.63}} \\

\midrule

\quad $Q_{\mathrm{act}}$
& 20 & 88.93 & 92.10 & 86.15 & 79.17 & 70.04 & 93.02 & 98.77 & 73.73
& 85.24 \deltapos{11.37} \\

\ourrow
\quad $Q_{\mathrm{grad}}$
& 20 & 92.14 & 92.50 & 91.00 & 85.49 & 73.64 & 93.62 & 98.66 & 74.17
& \textbf{87.65 \deltapos{13.78}} \\

\midrule

\grayrow
\multicolumn{11}{c}{
\textbf{ViT-L/14 $\rightarrow$ ViT-B/16 (Reduction)}
} \\
\midrule

\multicolumn{11}{l}{\textit{Reference checkpoints}} \\

\quad Source zero-shot
& -- & 72.51 & 67.73 & 58.55 & 72.54 & 66.37 & 93.02 & 86.66 & 73.55
& 73.87 \\

\quad Source fine-tuned
& -- & 99.14 & 98.30 & 99.20 & 96.98 & 85.70 & 94.93 & 99.76 & 82.62
& 94.58 \\

\quad Target zero-shot
& -- & 55.84 & 62.57 & 55.20 & 69.17 & 57.45 & 88.89 & 76.49 & 70.01
& 66.95 \deltazero \\

\midrule

\quad $Q_{\mathrm{act}}$
& 1 & 76.01 & 79.36 & 68.74 & 69.21 & 57.45 & 88.89 & 94.97 & 70.01
& 75.58 \deltapos{8.63} \\

\ourrow
\quad $Q_{\mathrm{grad}}$
& 1 & 82.67 & 75.03 & 64.50 & 71.62 & 58.27 & 88.84 & 93.82 & 70.01
& \textbf{75.60 \deltapos{8.65}} \\

\midrule

\quad $Q_{\mathrm{act}}$
& 2 & 76.63 & 84.44 & 70.53 & 69.98 & 57.45 & 88.89 & 95.89 & 70.01
& 76.73 \deltapos{9.78} \\

\ourrow
\quad $Q_{\mathrm{grad}}$
& 2 & 90.16 & 82.69 & 77.79 & 72.10 & 58.16 & 88.89 & 95.59 & 70.01
& \textbf{79.42 \deltapos{12.47}} \\

\midrule

\quad $Q_{\mathrm{act}}$
& 5 & 76.50 & 86.70 & 74.06 & 71.50 & 58.92 & 89.00 & 97.60 & 70.01
& 78.04 \deltapos{11.09} \\

\ourrow
\quad $Q_{\mathrm{grad}}$
& 5 & 88.93 & 84.07 & 78.84 & 76.93 & 62.53 & 89.01 & 96.37 & 69.97
& \textbf{80.83 \deltapos{13.88}} \\

\midrule

\quad $Q_{\mathrm{act}}$
& 10 & 79.30 & 88.76 & 75.92 & 72.36 & 60.22 & 88.89 & 98.23 & 70.01
& 79.21 \deltapos{12.26} \\

\ourrow
\quad $Q_{\mathrm{grad}}$
& 10 & 90.62 & 86.35 & 84.45 & 78.43 & 61.64 & 88.89 & 97.68 & 70.27
& \textbf{82.29 \deltapos{15.34}} \\

\midrule

\quad $Q_{\mathrm{act}}$
& 20 & 80.58 & 89.29 & 75.81 & 72.40 & 57.45 & 88.89 & 98.12 & 70.10
& 79.08 \deltapos{12.13} \\

\ourrow
\quad $Q_{\mathrm{grad}}$
& 20 & 91.23 & 90.23 & 86.20 & 78.94 & 61.94 & 89.03 & 98.22 & 70.46
& \textbf{83.28 \deltapos{16.33}} \\

\bottomrule
\end{tabular}
}
\end{table*}

\subsection{Resource Scaling}

\begin{table*}[t]
\caption{\textbf{Runtime and host-memory scaling with calibration budget.}
We compare \method{}, \theseus, and \bico across calibration budgets
$\mathcal{B}$.
Fit time is the median phase-total runtime per Vision8 task, including
activation collection, transformation, and transport.
Host memory is the maximum peak RSS across the eight Vision8 tasks and is
reported in decimal GB.
Relative time and relative host memory are normalized to \method{} at the
corresponding direction and calibration budget.
Baseline measurements at $\mathcal{B}=50$ are unavailable.}
\label{tab:supp_resource_scaling}
\centering
\setlength\tabcolsep{8pt}
\resizebox{\linewidth}{!}{%
\begin{tabular}{l c c c c c}
\toprule
\textbf{Method}
& $\mathcal{B}$
& \textbf{Fit time (s)}
& \textbf{Rel. time}
& \textbf{Host peak (GB)}
& \textbf{Rel. host} \\
\midrule

\grayrow
\multicolumn{6}{c}{
\textbf{ViT-B/16 $\rightarrow$ ViT-L/14 (Extension)}
} \\
\midrule

\quad \theseus
& 1 & 1023.53 & 76.7$\times$ & 73.06 & 4.74$\times$ \\
\quad \bico
& 1 & 933.46 & 69.9$\times$ & 89.42 & 5.80$\times$ \\
\ourrow
\quad \method{}
& 1 & \textbf{13.35} & \textbf{1$\times$} & \textbf{15.41} & \textbf{1$\times$} \\

\midrule

\quad \theseus
& 2 & 1075.37 & 60.5$\times$ & 71.42 & 4.44$\times$ \\
\quad \bico
& 2 & 1140.61 & 64.2$\times$ & 89.33 & 5.55$\times$ \\
\ourrow
\quad \method{}
& 2 & \textbf{17.77} & \textbf{1$\times$} & \textbf{16.10} & \textbf{1$\times$} \\

\midrule

\quad \theseus
& 5 & 1366.84 & 44.7$\times$ & 73.33 & 4.60$\times$ \\
\quad \bico
& 5 & 1323.15 & 43.2$\times$ & 90.43 & 5.67$\times$ \\
\ourrow
\quad \method{}
& 5 & \textbf{30.61} & \textbf{1$\times$} & \textbf{15.95} & \textbf{1$\times$} \\

\midrule

\quad \theseus
& 10 & 1919.15 & 36.2$\times$ & 73.19 & 4.53$\times$ \\
\quad \bico
& 10 & 1998.17 & 37.7$\times$ & 91.16 & 5.64$\times$ \\
\ourrow
\quad \method{}
& 10 & \textbf{53.07} & \textbf{1$\times$} & \textbf{16.16} & \textbf{1$\times$} \\

\midrule

\quad \theseus
& 20 & 3024.66 & 31.8$\times$ & 73.64 & 4.31$\times$ \\
\quad \bico
& 20 & 3205.04 & 33.7$\times$ & 90.84 & 5.31$\times$ \\
\ourrow
\quad \method{}
& 20 & \textbf{95.15} & \textbf{1$\times$} & \textbf{17.10} & \textbf{1$\times$} \\

\midrule

\ourrow
\quad \method{}
& 50 & \textbf{234.02} & \textbf{1$\times$} & \textbf{18.50} & \textbf{1$\times$} \\

\midrule

\grayrow
\multicolumn{6}{c}{
\textbf{ViT-L/14 $\rightarrow$ ViT-B/16 (Reduction)}
} \\
\midrule

\quad \theseus
& 1 & 426.41 & 90.5$\times$ & 38.83 & 2.02$\times$ \\
\quad \bico
& 1 & 423.76 & 90.0$\times$ & 48.00 & 2.49$\times$ \\
\ourrow
\quad \method{}
& 1 & \textbf{4.71} & \textbf{1$\times$} & \textbf{19.24} & \textbf{1$\times$} \\

\midrule

\quad \theseus
& 2 & 471.62 & 71.8$\times$ & 39.92 & 2.09$\times$ \\
\quad \bico
& 2 & 540.56 & 82.3$\times$ & 48.76 & 2.55$\times$ \\
\ourrow
\quad \method{}
& 2 & \textbf{6.57} & \textbf{1$\times$} & \textbf{19.14} & \textbf{1$\times$} \\

\midrule

\quad \theseus
& 5 & 600.22 & 49.6$\times$ & 39.87 & 2.02$\times$ \\
\quad \bico
& 5 & 619.89 & 51.2$\times$ & 50.16 & 2.54$\times$ \\
\ourrow
\quad \method{}
& 5 & \textbf{12.11} & \textbf{1$\times$} & \textbf{19.74} & \textbf{1$\times$} \\

\midrule

\quad \theseus
& 10 & 826.09 & 38.8$\times$ & 40.38 & 2.00$\times$ \\
\quad \bico
& 10 & 825.73 & 38.8$\times$ & 50.51 & 2.50$\times$ \\
\ourrow
\quad \method{}
& 10 & \textbf{21.29} & \textbf{1$\times$} & \textbf{20.22} & \textbf{1$\times$} \\

\midrule

\quad \theseus
& 20 & 1285.89 & 32.3$\times$ & 41.34 & 1.99$\times$ \\
\quad \bico
& 20 & 1370.81 & 34.5$\times$ & 50.06 & 2.41$\times$ \\
\ourrow
\quad \method{}
& 20 & \textbf{39.75} & \textbf{1$\times$} & \textbf{20.75} & \textbf{1$\times$} \\

\midrule

\ourrow
\quad \method{}
& 50 & \textbf{95.04} & \textbf{1$\times$} & \textbf{23.60} & \textbf{1$\times$} \\

\bottomrule
\end{tabular}
}
\end{table*}

\Cref{tab:supp_resource_scaling} shows that \method{}'s advantage is largest in the low-data regime: at one batch it is about 70--77$\times$ faster for extension and 90$\times$ faster for reduction. The relative gap narrows as calibration grows, but remains 32--35$\times$ at 20 batches. Absolute \method{} runtime rises from 13 to 234 seconds in extension and from 5 to 95 seconds in reduction between one and 50 batches. Host-memory growth is much slower, increasing by only about 20--23\% over the same range. At every jointly measured budget, \method{} also uses substantially less host memory than either baseline; the missing 50-batch baseline entries are therefore not used for a relative comparison.

\begin{table*}[t]
\caption{\textbf{Phase-level runtime decomposition.}
We decompose fit time into activation collection, transformation, and
transport. The upper block reports \method{} across calibration budgets,
while the lower block compares \method{}, \theseus, and \bico at
$\mathcal{B}=10$.
Values are median seconds per Vision8 task; evaluation and
merge-coefficient search are excluded.}
\label{tab:supp_phase_runtime}
\centering
\setlength\tabcolsep{7pt}

\resizebox{\linewidth}{!}{%
\begin{tabular}{l c c c c c}
\toprule
\textbf{Direction}
& $\mathcal{B}$
& \textbf{Collection (s)}
& \textbf{Transformation (s)}
& \textbf{Transport (s)}
& \textbf{Total fit (s)} \\
\midrule

\grayrow
\multicolumn{6}{c}{
\textbf{\method{} runtime scaling}
} \\
\midrule

\multirow{6}{*}{Extension}
& 1  & 0.80  & 12.53  & 0.03 & 13.35 \\
& 2  & 1.57  & 16.16  & 0.03 & 17.77 \\
& 5  & 3.89  & 26.69  & 0.03 & 30.61 \\
& 10 & 7.75  & 45.29  & 0.03 & 53.07 \\
& 20 & 15.48 & 79.63  & 0.03 & 95.15 \\
& 50 & 38.67 & 195.36 & 0.03 & 234.02 \\

\midrule

\multirow{6}{*}{Reduction}
& 1  & 0.99  & 3.71  & 0.01 & 4.71 \\
& 2  & 1.97  & 4.60  & 0.01 & 6.57 \\
& 5  & 4.90  & 7.20  & 0.01 & 12.11 \\
& 10 & 9.79  & 11.49 & 0.01 & 21.29 \\
& 20 & 19.59 & 20.14 & 0.01 & 39.75 \\
& 50 & 49.08 & 45.92 & 0.01 & 95.04 \\

\midrule

\grayrow
\multicolumn{6}{c}{
\textbf{Method comparison at $\mathcal{B}=10$}
} \\
\midrule

\multirow{3}{*}{Extension}
& \theseus
& 68.54 & 1845.84 & 5.00 & 1919.15 \\
& \bico
& 110.58 & 1881.09 & 4.96 & 1998.17 \\
\ourrow
& \method{}
& \textbf{7.75} & \textbf{45.29} & \textbf{0.03} & \textbf{53.07} \\

\midrule

\multirow{3}{*}{Reduction}
& \theseus
& 34.25 & 790.11 & 2.48 & 826.09 \\
& \bico
& 53.56 & 767.37 & 2.32 & 825.73 \\
\ourrow
& \method{}
& \textbf{9.79} & \textbf{11.49} & \textbf{0.01} & \textbf{21.29} \\

\bottomrule
\end{tabular}
}
\end{table*}

The decomposition in \cref{tab:supp_phase_runtime} locates the computational difference. For extension at $\mathcal{B}=10$, transformations account for 45 of \method{}'s 53 seconds, compared with roughly 1,850--1,880 seconds for the baselines. In reduction, \method{} divides its 21 seconds almost evenly between collection and transformation, while baseline transformations still require about 13 minutes. Applying the fitted update is negligible for \method{} in both directions. The measured speedup is therefore concentrated in the transformation stage rather than data collection or update application.

\subsection{Target Formulation and Regularization}

    \begin{table*}[t]
\caption{\textbf{Target formulation ablation.}
We compare the default change-target formulation of \method{} with an
endpoint-target variant at $\mathcal{B}=10$ calibration batches.
The default formulation transfers the source fine-tuning effect by regressing
the target update toward the fine-tuning-induced activation change,
$S^{\mathrm{ft}}_{\pi(j)} - S^{\mathrm{base}}_{\pi(j)}$ (mapped to target coordinates as $D_j$),
whereas the endpoint variant instead regresses the updated target directly
toward the absolute source fine-tuned activation $S^{\mathrm{ft}}_{\pi(j)}$.
Thus, the endpoint variant attempts to reproduce the source fine-tuned state
rather than transfer its functional change.
Thus, the endpoint variant attempts to reproduce the source fine-tuned state
rather than transfer its functional change.}
\label{tab:bank_endpoint_target_per_task}
\centering
\setlength\tabcolsep{4pt}
\resizebox{\linewidth}{!}{%
\begin{tabular}{l c c c c c c c c c}
\toprule
\textbf{Variant}
& EuroSAT
& SVHN
& GTSRB
& RESISC
& DTD
& Cars
& MNIST
& SUN
& \textbf{Avg ($\Delta$Acc)} \\
\midrule

\grayrow
\multicolumn{10}{c}{
\textbf{ViT-B/16 $\rightarrow$ ViT-L/14 (Extension)}
} \\
\midrule

\multicolumn{10}{l}{\textit{Reference checkpoints}} \\

\quad Source zero-shot
& 55.84 & 62.57 & 55.20 & 69.17 & 57.45 & 88.89 & 76.49 & 70.01
& 66.95 \\

\quad Source fine-tuned
& 98.68 & 97.91 & 99.04 & 96.12 & 82.27 & 93.24 & 99.73 & 78.55
& 93.19 \\

\quad Target zero-shot
& 72.51 & 67.73 & 58.55 & 72.54 & 66.37 & 93.02 & 86.66 & 73.55
& 73.87 \deltazero \\

\midrule

\ourrow
\quad \method{} (change target)
& 87.98 & 91.02 & 83.26 & 78.87 & 68.68 & 93.02 & 98.50 & 73.73
& \textbf{84.38 \deltapos{10.52}} \\

\quad \method{} (endpoint target)
& 77.65 & 74.80 & 61.20 & 72.54 & 66.37 & 93.02 & 90.77 & 73.55
& 76.24 \deltapos{2.37} \\

\midrule

\grayrow
\multicolumn{10}{c}{
\textbf{ViT-L/14 $\rightarrow$ ViT-B/16 (Reduction)}
} \\
\midrule

\multicolumn{10}{l}{\textit{Reference checkpoints}} \\

\quad Source zero-shot
& 72.51 & 67.73 & 58.55 & 72.54 & 66.37 & 93.02 & 86.66 & 73.55
& 73.87 \\

\quad Source fine-tuned
& 99.14 & 98.30 & 99.20 & 96.98 & 85.70 & 94.93 & 99.76 & 82.62
& 94.58 \\

\quad Target zero-shot
& 55.84 & 62.57 & 55.20 & 69.17 & 57.45 & 88.89 & 76.49 & 70.01
& 66.95 \deltazero \\

\midrule

\ourrow
\quad \method{} (change target)
& 79.30 & 88.76 & 75.92 & 72.36 & 60.22 & 88.89 & 98.23 & 70.01
& \textbf{79.21 \deltapos{12.26}} \\

\quad \method{} (endpoint target)
& 63.54 & 84.68 & 75.98 & 70.49 & 57.45 & 88.89 & 93.34 & 70.01
& 75.55 \deltapos{8.60} \\

\bottomrule
\end{tabular}
}
\end{table*}

\Cref{tab:bank_endpoint_target_per_task} gives the per-task version of the target ablation. Regressing toward the absolute source fine-tuned endpoint remains better than the target base on average, particularly in reduction, so the endpoint contains some transferable information. It is nevertheless consistently weaker than matching the source-induced change: the gap is 8.14 points in extension and 3.66 points in reduction. The larger extension penalty is driven by broad losses on EuroSAT, SVHN, GTSRB, and RESISC45. These results support treating the target base as the reference state and importing only the task-specific displacement, rather than asking a heterogeneous target to reproduce source features that are unrelated to fine-tuning.

\begin{table*}[t]
\caption{\textbf{Ridge selection.}
We compare the empirical-Bayes ridge used by \method{} with fixed
trace-scaled coefficients $\rho\in\{0.001,0.01,0.1,1,10\}$. Entries are Vision8 test
accuracy (\%), with $\alpha$ selected per task on validation.
Bold averages indicate the best ridge choice within each transfer
direction.}
\label{tab:supp_vision8_ridge_choice}
\centering
\setlength\tabcolsep{4pt}
\resizebox{\linewidth}{!}{%
\begin{tabular}{l c c c c c c c c c}
\toprule
\textbf{Ridge}
& EuroSAT & SVHN & GTSRB & RESISC & DTD & Cars & MNIST & SUN
& \textbf{Avg} \\
\midrule

\grayrow
\multicolumn{10}{c}{\textbf{ViT-B/16 $\rightarrow$ ViT-L/14 (Extension)}} \\
\midrule

\multicolumn{10}{l}{\textit{Reference checkpoint}} \\

\quad Target zero-shot
& 72.51 & 67.73 & 58.55 & 72.54 & 66.37 & 93.02 & 86.66 & 73.55
& 73.87 \\

\midrule

\quad Fixed $\rho=0.001$
& 88.23 & 91.47 & 82.86 & 78.59 & 69.39 & 93.02 & 98.51 & 73.73
& 84.47 \\

\quad Fixed $\rho=0.01$
& 88.15 & 91.41 & 83.13 & 78.84 & 69.80 & 93.02 & 98.50 & 73.71
& \textbf{84.57} \\

\quad Fixed $\rho=0.1$
& 88.27 & 90.71 & 82.98 & 78.57 & 67.49 & 93.02 & 98.51 & 73.71
& 84.16 \\

\quad Fixed $\rho=1$
& 87.16 & 89.01 & 80.24 & 77.51 & 69.33 & 93.02 & 98.11 & 73.61
& 83.50 \\

\quad Fixed $\rho=10$
& 85.68 & 85.09 & 74.80 & 75.84 & 67.26 & 93.02 & 96.96 & 73.62
& 81.53 \\

\ourrow
\quad Empirical Bayes
& 88.27 & 91.12 & 83.43 & 78.87 & 68.68 & 93.02 & 98.50 & 73.73
& 84.45 \\

\midrule

\grayrow
\multicolumn{10}{c}{\textbf{ViT-L/14 $\rightarrow$ ViT-B/16 (Reduction)}} \\
\midrule

\multicolumn{10}{l}{\textit{Reference checkpoint}} \\

\quad Target zero-shot
& 55.84 & 62.57 & 55.20 & 69.17 & 57.45 & 88.89 & 76.49 & 70.01
& 66.95 \\

\midrule

\quad Fixed $\rho=0.001$
& 79.79 & 88.71 & 75.77 & 71.38 & 59.81 & 88.97 & 98.37 & 70.01
& 79.10 \\

\quad Fixed $\rho=0.01$
& 79.92 & 88.84 & 75.97 & 71.39 & 59.63 & 89.01 & 98.09 & 70.01
& \textbf{79.11} \\

\quad Fixed $\rho=0.1$
& 79.18 & 88.62 & 75.68 & 72.17 & 57.92 & 88.89 & 98.01 & 70.06
& 78.82 \\

\quad Fixed $\rho=1$
& 78.07 & 87.28 & 73.06 & 71.16 & 57.45 & 89.01 & 97.16 & 70.01
& 77.90 \\

\quad Fixed $\rho=10$
& 75.10 & 79.66 & 69.83 & 70.19 & 57.45 & 88.90 & 91.16 & 70.01
& 75.29 \\

\ourrow
\quad Empirical Bayes
& 78.52 & 88.28 & 75.92 & 71.48 & 57.45 & 88.89 & 98.23 & 70.01
& 78.60 \\

\bottomrule
\end{tabular}
}
\end{table*}

\Cref{tab:supp_vision8_ridge_choice} shows a broad useful range for the target-update regularizer. Fixed trace-scaled coefficients from $0.001$ to $0.1$ produce similar averages, while excessive shrinkage at $\rho=10$ removes 3.04 points in extension and 3.82 points in reduction relative to the best fixed choice. The empirical-Bayes rule is within 0.12 points of the best fixed coefficient in extension and 0.51 in reduction without validation-time tuning. These results support the automatic ridge scale as a practical default, with little loss relative to the best tested fixed coefficients.

\end{document}